\documentclass[runningheads]{llncs}
\usepackage{graphicx}
\usepackage{comment}
\usepackage{amsmath,amssymb} %
\usepackage{color}
\usepackage{booktabs}
\usepackage{multirow}
\usepackage{enumitem}
\usepackage{siunitx}
\usepackage{amsmath}
\usepackage{xspace}
\usepackage[table,xcdraw]{xcolor}
\usepackage[font=small,labelfont=bf]{caption}
\usepackage{bm}
\usepackage{subcaption}
\usepackage{cite}
\usepackage{hyperref}
\hypersetup{
    colorlinks=true,
    citecolor=blue,
    linkcolor=blue
}

\usepackage[width=122mm,left=12mm,paperwidth=146mm,height=193mm,top=12mm,paperheight=217mm]{geometry}

\newcommand{\etal}{et al.}
\newcommand{\ie}{i.e.}
\newcommand{\eg}{e.g.}

\DeclareMathOperator*{\argmin}{arg\,min}
\newcommand{\amplitude}{\alpha}
\newcommand{\phase}{\beta}
\newcommand{\dcoffset}{\gamma}
\newcommand{\sinogram}{\bm{\tau}_{\text{circ}}}

\newcommand{\source}{\mathbf{x}''}
\newcommand{\sensor}{\mathbf{x}'}
\newcommand{\voxel}{\mathbf{x}}
\newcommand{\CCNLOS}{$\text{C}^2\text{NLOS}$\xspace}

\begin{document}
\pagestyle{headings}
\mainmatter
\def\ECCVSubNumber{177}  %

\title{Efficient Non-Line-of-Sight Imaging \\ from Transient Sinograms}

\titlerunning{Efficient Non-Line-of-Sight Imaging from Transient Sinograms}
\author{
Mariko Isogawa \and
Dorian Chan \and
Ye Yuan \and
Kris Kitani \and
Matthew O'Toole
}

\authorrunning{M. Isogawa et al.}

\institute{Carnegie Mellon University}

\maketitle

\begin{abstract}
Non-line-of-sight (NLOS) imaging techniques use light that diffusely reflects off of visible surfaces (\eg, walls) to see around corners.  One approach involves using pulsed lasers and ultrafast sensors to measure the travel time of multiply scattered light. Unlike existing NLOS techniques that generally require densely raster scanning points across the entirety of a relay wall, we explore a more efficient form of NLOS scanning that reduces both acquisition times and computational requirements.  We propose a circular and confocal non-line-of-sight (\CCNLOS) scan that involves illuminating and imaging a common point, and scanning this point in a circular path along a wall.  We observe that (1) these \CCNLOS measurements consist of a superposition of sinusoids, which we refer to as a transient sinogram, (2) there exists computationally efficient reconstruction procedures that transform these sinusoidal measurements into 3D positions of hidden scatterers or NLOS images of hidden objects, and (3) despite operating on an order of magnitude fewer measurements than previous approaches, these \CCNLOS scans provide sufficient information about the hidden scene to solve these different NLOS imaging tasks.  We show results from both simulated and real \CCNLOS scans.\footnote{Project page: \href{https://marikoisogawa.github.io/project/c2nlos}{\texttt{https://marikoisogawa.github.io/project/c2nlos}}}
\end{abstract}

\begin{figure}[t]
\begin{center}
\includegraphics[width=1.0\textwidth]{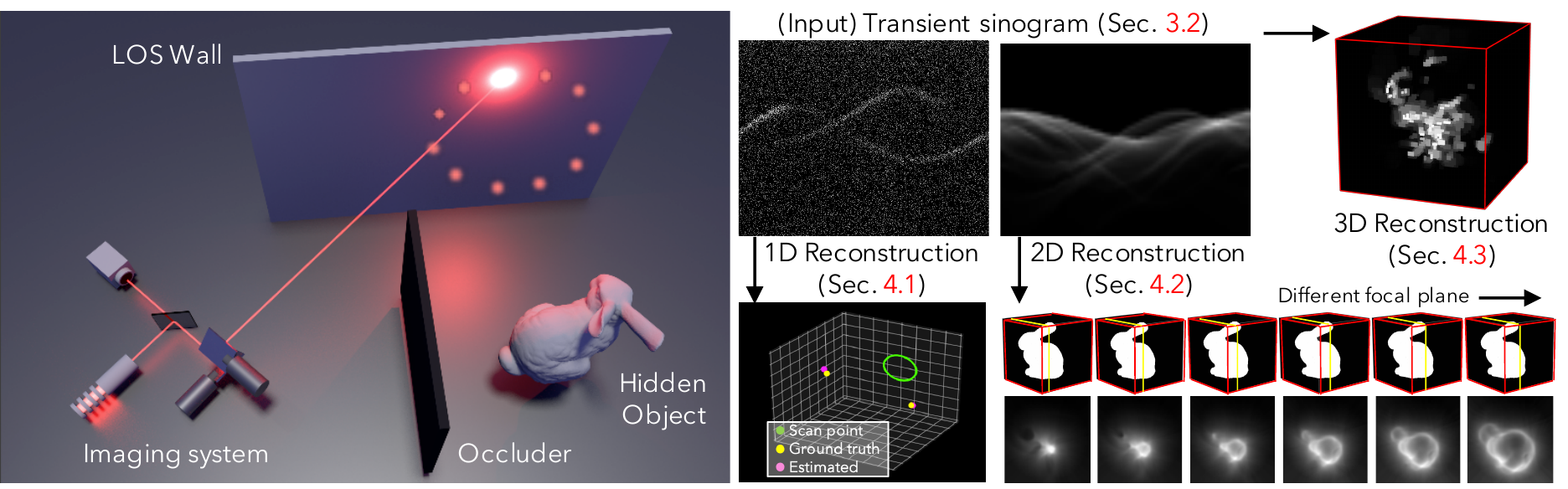}
\end{center}
\caption{A circular and confocal non-line-of-sight (\CCNLOS) system scans points along a circular path on a relay wall. Exploiting the sinusoidal properties of \CCNLOS measurements, a circular scan of a wall is sufficient to reconstruct images.  \CCNLOS operates on far fewer measurements than existing NLOS techniques.}
\label{fig:teaser}
\end{figure}

\section{Introduction}

The ability to image objects hidden outside of a camera's field of view has many potential applications~\cite{maeda2019recent}, including autonomous driving, search and rescue, and remote imaging. Over the last decade, many different technologies have been used for non-line-of-sight (NLOS) imaging, including transient imaging~\cite{Ahn2019,Buttafava2015,Chan2017,Gupta2012,Heide2014,Heide2019,Isogawa_2020_CVPR,Kirmani2009,LaManna2019,Lindell2019,metzler_arxiv2019,o2018confocal,SNLOS,redo2016terahertz,Tsai2019,Velten2012,Xin2019}, conventional cameras~\cite{Bouman2017,chandran_bmvc2015,Klein2016,Klein2017,Saunders2019,Tancik2018}, WiFi or radio frequency measurements~\cite{Adib2015,Li2019}, thermal imaging~\cite{Maeda2019}, and even audio-based techniques~\cite{lindell2019acoustic}.  Transient imaging refers to measuring a scene's temporal response to a pulse of light, and is one of the more successful approaches to reconstructing high-quality 3D shape of hidden scenes.

 NLOS imaging techniques are fundamentally dependent on the spatial scanning patterns they utilize. Initially, methods used exhaustive measurements of 5D transients~\cite{Ahn2019,Gupta2012,Kirmani2009,Velten2012}, requiring explicit scanning of both virtual sources and sensors on a line-of-sight (LOS) wall. To mitigate these issues, alternative approaches proposed co-locating the source and sensor points~\cite{o2018confocal}, reducing the dimensionality of the required scanning to just two spatial dimensions which significantly expedites computation. However, even these techniques still require a full raster scan of a wall, which is limited to \SI{2}{\hertz} to \SI{4}{\hertz} for state-of-the-art NLOS systems~\cite{Lindell2019}---too slow for real-time capture. 

All of these previous NLOS techniques motivate the following two-fold question.  First, what is the dimensionality of the smallest set of measurements that is sufficient for reconstructing a NLOS image? And second, among all measurement sets of this size, which ones lend themselves to efficient reconstruction algorithms?  Answering these questions involves many complicated considerations, including the need to define the exact reconstruction problem we are solving.

While we do not provide definitive answers to these questions in this paper, we take first steps towards addressing them. In particular, we identify the subset of measurements produced by a circular and confocal non-line-of-sight (\CCNLOS) scan, which yields powerful properties that facilitate fast reconstructions.  As shown in Fig.~\ref{fig:teaser}, \CCNLOS scanning involves sampling points that form a circle on a visible surface, reducing the dimensionality of transient measurements under this regime to just 2 dimensions.  Our key observation is that NLOS images can be obtained with far fewer measurements than previously expected or demonstrated by existing NLOS systems and reconstruction techniques.  With off-the-shelf large beam scanning galvo systems (\eg, Thorlabs GVS012), circular scanning is also fast and potentially supports real-time NLOS tasks at \SI{130}{\hertz}. 

In addition to having smaller dimensionality and being efficient to acquire, \CCNLOS measurements satisfy the requirements set out above, sufficiency and computational efficiency, for two important NLOS reconstruction problems. The first problem is localizing a discrete number of small objects (``scatterers").  We show that \CCNLOS measurements are sufficient for this task, and enable recovery of the unknown locations through a straightforward Hough voting procedure~\cite{houghvote}. The second problem is reconstructing a single planar object. For small planar objects, we show that this problem can be reduced to one equivalent to computed tomography, and therefore can be solved using techniques developed for that task such as the inverse Radon transform~\cite{kak2002principles}. Both results rely on a theoretical analysis that shows that the transient measurements from \CCNLOS scanning are a superposition of sinusoids (referred to as a \emph{transient sinogram}) with different amplitudes, phases, and offsets.  A one-to-one mapping directly relates the parameters of these sinusoids to the 3D position of the hidden scatterers.

Motivated by the above results, we also empirically investigate two related problems. We show that accurate 2D images of large planar scenes can be obtained by solving a simple linear least squares problem based on \CCNLOS measurements. Furthermore, we demonstrate that approximate 3D reconstructions of the NLOS scene can be efficiently recovered from \CCNLOS measurements.

To summarize, our contributions are the following:
 (i) we provide a theoretical analysis of our proposed \CCNLOS scanning procedure which shows that the measurements consist of a superposition of sinusoids, producing a transient sinogram; (ii) we propose efficient reconstruction procedures that build on these sinusoidal properties to localize hidden objects and reconstruct NLOS images; and (iii) we show that the \CCNLOS measurements are sufficient for reconstructing 1D, 2D, and 3D NLOS images from both simulated and real transients, while using far fewer measurements than existing methods.

It should be stated up front that our reconstruction quality is strictly worse than conventional methods that make use of a larger set of transient measurements and longer capture times. In contrast to past NLOS works where the objective is to improve reconstruction quality, our key objective is to show that \textbf{a circular scan of transients is sufficient to reconstruct a NLOS image}.

\section{Overview of Transient NLOS Imaging}

NLOS imaging has received significant attention, with solutions that operate on a wide variety of different principles. However, a common approach to NLOS imaging involves using transient sources and sensors that operate on visible or near-IR light; we refer to these as transient NLOS imaging systems.

\CCNLOS imaging is a transient-based technique which shares many similarities to previous work in confocal NLOS imaging~\cite{o2018confocal}.  We therefore review the general NLOS image formation model, followed by confocal NLOS imaging.

\subsubsection{General NLOS Imaging~\cite{Velten2012}.}

\begin{figure}[t]
\begin{center}
\includegraphics[width=1.0\textwidth]{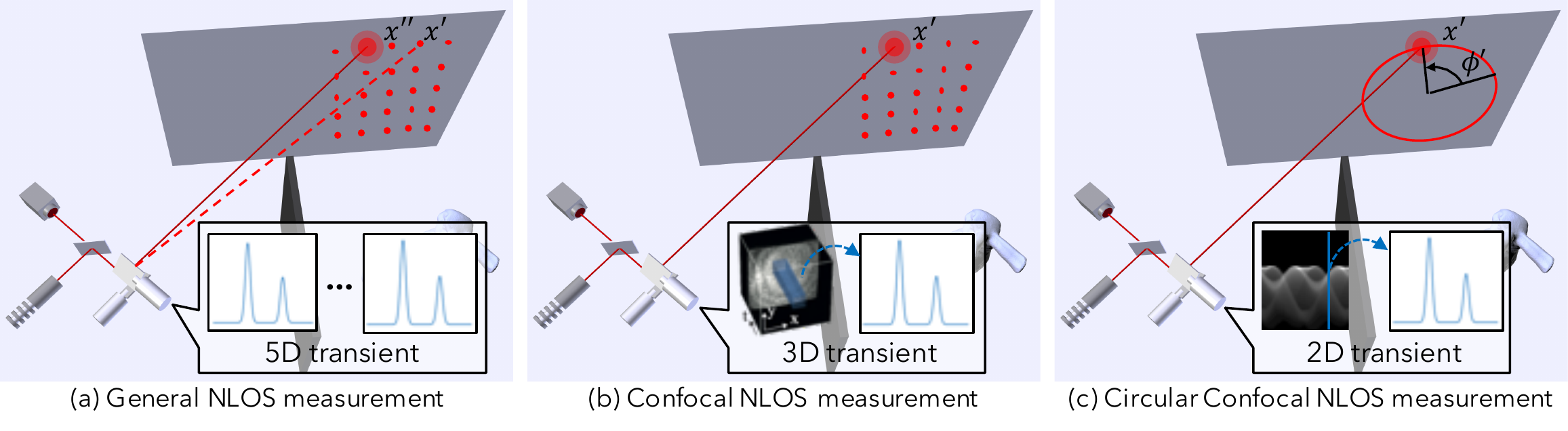}
\end{center}
\caption{Illustration of transient NLOS scans. A pulsed laser illuminates a point $\source$, while a transient sensor images a point $\sensor$ on a LOS wall. By changing $\source$ and $\sensor$, we obtain a diverse set of transients that can help identify the position, shape, and appearance of objects hidden from sight.
\textbf{(a)} Conventional NLOS imaging scans several combinations of light source positions $\source$ and detector positions $\sensor$ on the wall to obtain a 5D transient measurement. 
\textbf{(b)} Confocal NLOS imaging illuminates and images the same point, \ie, $\source = \sensor$, producing a 3D transient measurement. 
\textbf{(c)} \CCNLOS imaging proposes confocally scanning only those points that lie on a circle, yielding a 2D transient measurement. We propose a transformation that reduces this transient image into a sum of sinusoids, called a \emph{transient sinogram}, where the amplitude, phase, and offset of each sinusoid corresponds to the position of a hidden scatterer.}
\label{fig:NLOS_explain}
\end{figure}

In Fig.~\ref{fig:NLOS_explain}(a), a laser sends a pulse of light towards a 3D point $\source$ on a visible wall, and the light diffusely scatters from that point at time $t=0$.  The scattered light illuminates the objects hidden around a corner, and a fraction of that light reflects back towards the wall in response.  A transient sensor (\eg, a SPAD) then measures the temporal response at a point $\sensor$ on the wall, also known as a transient measurement~\cite{O'Toole2017}. The transient measurement, $\tau(\sensor,\source,t)$, represents the amount of light detected at point $\sensor$ at time $t$, given illumination from point $\source$ at time $t = 0$.  For simplicity, we ignore the travel time between the system to the wall itself, which can be accounted for given the wall's geometry relative to the position of the laser and sensor.

The standard image formation model for NLOS imaging is
\begin{equation}
    \tau(\sensor, \source,t) = \iiint_{\Omega} \rho(\voxel) ~\frac{\delta\left(\|\voxel-\sensor\| + \|\voxel-\source\| - tc\right)}{\|\voxel-\sensor\|^2\|\voxel-\source\|^2}~d\voxel\label{eq:nlos_model},
\end{equation}
where the function $\rho(\voxel)$ represents the albedo of objects at every point $\voxel$, and $c = 3\times10^8~\text{m/s}$ is the speed of light.  The expression inside the Dirac delta $\delta(\cdot)$ relates the distance light travels through the hidden volume to its time of flight.  The denominator accounts for the decrease in the intensity of light as a function of distance traveled, as given by the inverse square law.

This image formation model makes three underlying assumptions: (i) it only models three-bounce light paths, (ii) the model ignores the effect of a material's reflectance function and surface orientation, and (iii) the model assumes no occlusions within the hidden volume.

Equation~\eqref{eq:nlos_model} can be discretized into a linear system of equations
\begin{equation}
    \bm{\tau} = \mathbf{A} \bm{\rho},\label{eq:nlos_discrete_model}
\end{equation}
where $\bm{\tau}$ and $\bm{\rho}$ are discretized and vectorized representations of the measurements and volume, respectively.  Recovering the hidden scene's geometry involves solving the linear system in Equation~\eqref{eq:nlos_discrete_model}.  Unfortunately, the matrix $\mathbf{A}$ can be extremely large in practice.  In general, the matrix maps a 3D volume $\bm{\rho}$ to a 5D transient measurement $\bm{\tau}$ (4D spatial + 1D temporal).  In this case, the matrix is far too large to construct, store, and invert directly.

As a result, many works have explored different sampling patterns that reduce the size of the measurements and simplify the reconstruction procedure.  Certain approaches simply fix the light source and scan the sensor (or vice versa), producing 2D spatial measurements \cite{Buttafava2015,Heide2014}.  SNLOS~\cite{SNLOS} temporally focuses the light reflecting off of a single voxel, by simultaneously illuminating and imaging the wall over an ellipse.  To scan a 2D or 3D set of voxels, ellipses of different shapes and sizes are used.  Keyhole NLOS imaging \cite{metzler_arxiv2019} illuminates and detects light at a single point on the LOS wall, and relies on the motion of the hidden object to produce measurements for NLOS imaging. Confocal NLOS imaging~\cite{Lindell2019,o2018confocal} scans the source and sensor together, and is described next in more detail.

\subsubsection{Confocal NLOS Imaging~\cite{o2018confocal}.} \label{subsec:confocal}

Confocal NLOS imaging (Fig.~\ref{fig:NLOS_explain}(b)) co-locates the source and sensor by setting $\sensor = \source$, and samples a regular 2D grid of points on the wall.  This sampling strategy has a number of practical advantages.  First, it simplifies the NLOS calibration process, since the shape of the wall is given by direct reflections.  Second, confocal scans capture light from retroreflective objects, which helps to enable NLOS imaging at interactive rates~\cite{Lindell2019}. Finally, there exist computationally and memory efficient algorithms for recovering hidden volumes from confocal scans without explicit construction of matrix $\mathbf{A}$.
 
When co-locating the source and detector, Equation~\eqref{eq:nlos_model} reduces to
\begin{equation}
    \tau(\mathbf{x}',t) = \iiint_{\Omega} \rho(\mathbf{x}) ~\frac{\delta\left(2\|\mathbf{x}'-\mathbf{x}\| - tc\right)}{{\|\mathbf{x}'-\mathbf{x}\|^4}}~d\mathbf{x}.
\end{equation}
As discussed by O'Toole~\etal~\cite{o2018confocal}, a change of variables $v = (tc/2)^2$ produces
\begin{equation}
    \tilde{\tau}(\sensor,v) \equiv v^{\frac{3}{2}}~\tau(\mathbf{x}',\tfrac{2}{c} \sqrt{v}) = \iiint_{\Omega} \rho(\mathbf{x})~\delta\left(\|\mathbf{x}'-\mathbf{x}\|^2 - v\right)~d\mathbf{x}.
    \label{eq:confocal}
\end{equation}
When the relay wall is planar (\ie, $z' = 0$), the 3D spatio-temporal response $\tilde{\tau}(\sensor,v)$ of a scatterer becomes shift-invariant with respect to its 3D position $\mathbf{x}$.  Equation~\eqref{eq:confocal} can then be expressed as a simple 3D convolution, which can be efficiently evaluated using a fast Fourier transform.  The inverse problem involves a simple 3D deconvolution procedure called the light cone transform (LCT).

\section{The Geometry of Circular and Confocal Scanning}

\begin{figure}[t]
\begin{center}
\includegraphics[width=1.0\textwidth]{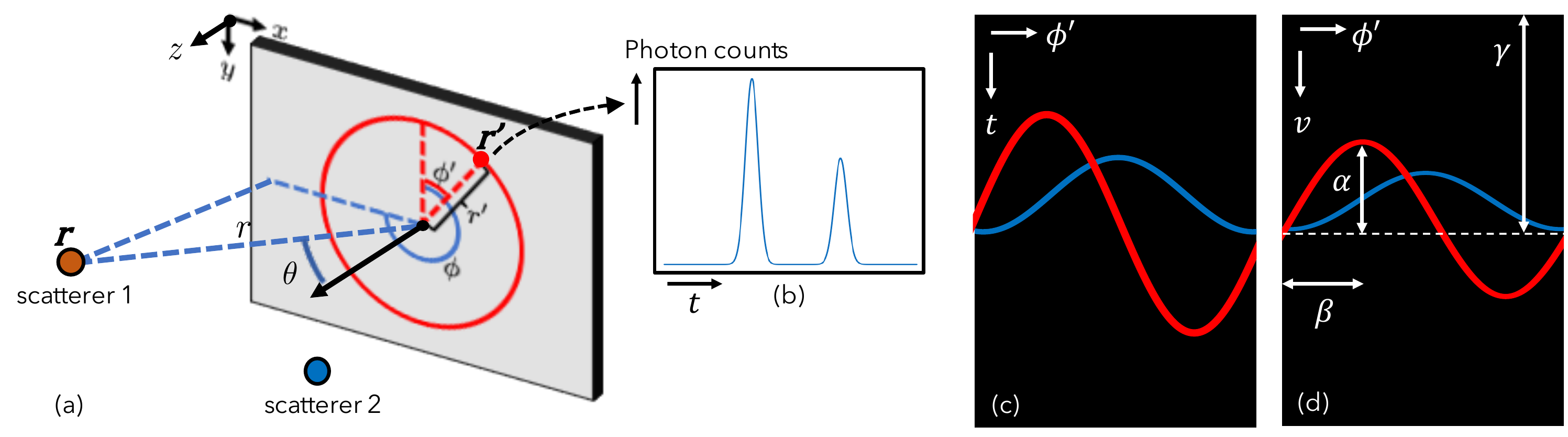}
\end{center}
\caption{Geometry of a \CCNLOS scan for individual scatterers.  \textbf{(a)} The system confocally scans the red circle of radius $r'$ one point at a time to image hidden objects.  \textbf{(b)} Each point produces a transient, \ie, the temporal response to a pulse of light.  This signal represents the travel time from a point on the wall, to the scatterers, and back again.  \textbf{(c)} Scanning different points on the circle produces a collection of transients.  Note that the signals represented here are only approximately sinusoidal.  \textbf{(d)} By resampling the transients through a change of variables $v = (tc/2)^2$ (as explained in Eq.~\eqref{eq:confocal}), we obtain a transient sinogram.  Every scatterer produces a unique sinusoid with a specific amplitude $\alpha$, phase $\beta$, and offset $\gamma$.  The parameters of these sinusoids are directly related to the spherical coordinates of the scatterers; see Eq.~\eqref{eq:param_to_pos}.}
\label{fig:sinusoid_scatter}
\end{figure}

While previous approaches have successfully reduced both capture and reconstruction times, the scanning paths required by these techniques inherently restrict scanning speeds on current hardware. Typical NLOS imaging systems, such as the one developed by Lindell~\etal~\cite{Lindell2019}, use a pair of large galvo mirrors to raster scan the wall.  The mirrors can only be driven up to a maximum of \SI{65}{\hertz} for a square wave pattern, capping scanning to just $130$ lines per second. As a result, even the 2D grids utilized by confocal approaches~\cite{Lindell2019,o2018confocal} are limited to just a few hertz (e.g., $32 \times 32$ at \SI{4}{\hertz}, or $64 \times 64$ at \SI{2}{\hertz}), impractical for dynamic scenes. Higher dimensional non-confocal measurements are even slower. Although smaller mirrors enable higher-frequency modulation (\eg, MEMS mirrors operate at kHz rates), this would greatly reduce the light efficiency of the system, lowering the quality of the output measurement.

This fundamental mechanical limitation motivates the following question: \emph{can we further reduce the scanning path to just a single dimension, while still capturing useful information about the hidden scene?} We analyze the case of a circular and confocal scan (see Fig.~\ref{fig:NLOS_explain}(c)). Such a sinusoidal pattern could easily be captured at 130 Hz under current galvo-mirror systems---a typical NLOS setup can capture an entire 1D circular scan in the time it takes to capture a single row of a 2D grid scan. At the same time, these \CCNLOS scans encode significant information about the hidden scene. We investigate their properties in further detail in the rest of this section.

\subsection{Equation~\eqref{eq:confocal} in Spherical Coordinates}
\label{sec:confocal_spherical}

We start by analyzing the form of Equation~\eqref{eq:confocal} when expressed in spherical coordinates. As shown in Fig.~\ref{fig:sinusoid_scatter}(a),  we express the position of voxels in the hidden scene and scanning positions on the wall as $\mathbf{r} = (r, \theta, \phi)$ and $\mathbf{r}' = (r', \theta', \phi')$ respectively, where
\begin{eqnarray}
    x = r \sin(\theta) \cos(\phi), 
    ~~y = r \sin(\theta) \sin(\phi),
    ~~z = r \cos(\theta),\label{eq:sphericalcoordinates}
\end{eqnarray}
for an azimuth angle $0 \leq \phi \leq 2\pi$, a zenith angle $0 \leq \theta \leq \pi$, and a radius $r \geq 0$.

Through a change of variables, we can rewrite the confocal NLOS image formation model of Equation~\eqref{eq:confocal} as
\begin{equation}
    \tilde{\tau}(\mathbf{r}',v) \equiv v^{\frac{3}{2}}~\tau(\mathbf{r}',\tfrac{2}{c} \sqrt{v}) = \iiint_{\Omega} \rho(\mathbf{r})~\delta\left(d(\mathbf{r}',\mathbf{r})^2 - v\right)~r^2\sin(\theta)~dr~d\theta~d\phi,
\end{equation}
where the distance function $d(\cdot,\cdot)$ expressed in spherical coordinates is
\begin{equation}
    v(\mathbf{r}') \equiv d(\mathbf{r}',\mathbf{r})^2 = r^2 + r'^2 - 2rr'\left(\sin(\theta)\sin(\theta')\cos(\phi-\phi')+\cos(\theta)\cos(\theta')\right).\label{eq:circular_diff}
\end{equation}
We restrict scatterers to be on one side of the wall, by setting $\theta \leq \pi/2$.

\subsection{Transient Sinograms}
\label{sec:circular_nlos}

We assume that our \CCNLOS scans points along a wall where $\theta' = \pi/2$, the points on the wall are on a circle of fixed radius $r'$, and the center of the circle is the origin $\mathbf{0}$.  By applying these assumptions to Equation~\eqref{eq:circular_diff}, we get
\begin{eqnarray}
v(\phi') = r^2 + r'^2 - 2rr'\sin(\theta)\cos(\phi-\phi') = \dcoffset - \amplitude \cos(\phase - \phi'),
\end{eqnarray}
where
\begin{eqnarray}
    \amplitude = 2rr'\sin(\theta), \quad\phase = \phi, \quad\dcoffset = r^2 + r'^2.
    \label{eq:pos_to_param}
\end{eqnarray}
Here, $\alpha$, $\beta$, and $\gamma$ represent the amplitude, phase, and offset of a sinusoid.  Therefore, after resampling the transient measurements (Fig.~\ref{fig:sinusoid_scatter}(c)) with the substitution $v = (tc/2)^2$, the transient measurement resulting from a \CCNLOS scan becomes a 2D image representing a superposition of different sinusoids, where each sinusoid represents a different point in the hidden space~(see Fig.~\ref{fig:sinusoid_scatter}(d)).  We therefore refer to the corresponding measurement as a \emph{transient sinogram} $\bm{\tau}_{\text{circ}}$.

Consider scatterers of the form $\mathbf{r} = (r,0,0)$ for all $r \geq 0$.  These scatterers produce a sinusoid with zero amplitude, because all points on the circle are equidistant to the scatterer.  As we change the zenith angle $\mathbf{r} = (r,\theta,0)$ for $0 \leq \theta \leq \pi / 2$, the amplitude of the sinusoid also increases up to a maximum of $2rr'$ when scatterers are adjacent to the wall.  Finally, introducing an azimuth angle $\mathbf{r} = (r,\theta,\phi)$ for $0 \leq \phi \leq 2\pi$ produces a phase shift of the sinusoid.

After identifying the amplitude, phase, and offset of each scatterer's sinusoid, inverting the expression in Equation~\eqref{eq:pos_to_param} recovers the scatterer's position:
\begin{eqnarray}
    r = \sqrt{\gamma -r'^2}, \quad \theta = \arcsin{\left(\frac{\alpha}{2rr'}\right)}, \quad \phi = \beta\label{eq:param_to_pos}.
\end{eqnarray}
This expression becomes useful when estimating the positions of one or a handful of scatterers around a corner, as discussed in Section~\ref{sec:object_pos_estimation}.

In addition to computing the position of scatterers, another NLOS objective is to reconstruct images of the hidden scene.  Consider an object that lies on the surface of a sphere of known radius $r$ and centered about point $\mathbf{0}$ on the wall.  This scenario occurs when sufficiently-small planar objects are oriented towards the origin $\mathbf{0}$.  (For planar objects tilted away from the origin, one can scan a different circle centered about another point on the wall.)  According to Equation~\eqref{eq:pos_to_param}, the measurement consists of a combination of sinusoids, all of which have identical offsets.  This simplifies the measurement into a conventional sinogram image, the same type of measurement used in computed tomography (CT)~\cite{kak2002principles}. We exploit this property in Section~\ref{sec:inverse_radon_method} to recover 2D images of the hidden scene.

\section{Reconstructing Images from Transient Sinograms}

\subsection{1D Reconstruction: Estimating 3D Positions}
\label{sec:object_pos_estimation}

Given a transient sinogram $\sinogram$, our first goal is to recover the 3D position of an object located at $\mathbf{x} = (x,y,z)$ by estimating its corresponding sinusoid parameters $\amplitude, \phase$, and $\dcoffset$, as described in Equation~\eqref{eq:param_to_pos}.  The challenge is to perform this operation both accurately and robustly, \eg, in the presence of sensor noise.

We propose a convolutional approach to the Hough transform for fixed-period sinusoids, based loosely on~\cite{sinusoid_hough}.
First, we generate a 2D Hough kernel representing a sinusoid of a given amplitude.  Through a FFT-based convolution between this kernel and the transient sinogram, we obtain a parameter-space image that produces large responses in areas where the kernel aligns well with the sinusoid in the transient sinogram.  Second, we repeat this procedure multiple times for kernels representing sinusoids with different amplitudes, producing a three dimensional parameter-space volume.  The location of the voxels with the highest values in the volume represent the parameters of the sinusoids in the transient sinogram. Using these sinusoids, we can then recover the scatterers in the hidden scene by applying Equation~\eqref{eq:param_to_pos}. We illustrate this process in Fig.~\ref{fig:1D_tracking_result}(left).

\subsection{2D Reconstructions}
\label{sec:2d_iradon_explain}

Consider the scenario where the hidden scene can be approximately modelled as a single planar object.  We propose two ways to reconstruct a 2D image of this object.  First, the inverse Radon transform is an integral transform used to solve the CT reconstruction problem; it is therefore possible to directly apply the inverse Radon transform technique on transient sinograms to recover a 2D image. This assumes that the planar patch is small, and tangent to the surface of a sphere of radius $r$, with the same center as the \CCNLOS scan. Second, since it is shown transient sinograms preserve information about the hidden scene and the measurements are much smaller when compared to conventional NLOS scans, it becomes computationally feasible to explicitly construct a matrix $\mathbf{A}$ that directly maps points from a hidden 2D plane to the \CCNLOS measurements, and solve a discrete linear system (\ie, Equation~\eqref{eq:nlos_discrete_model}) directly.

\subsubsection{Inverse Radon Reconstruction} 
\label{sec:inverse_radon_method}
\begin{figure}[t]
\begin{center}
\includegraphics[width=1.0\textwidth]{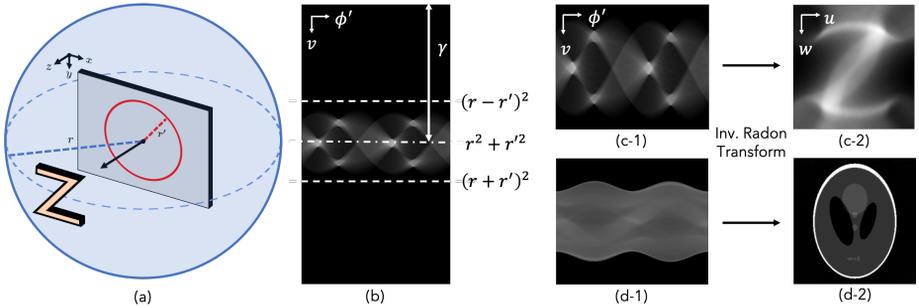}
\end{center}
\caption{Inverse Radon reconstruction based 2D imaging. \textbf{(a)} Suppose the hidden Z is a planar object that approximately lies on the surface of a sphere of radius $r$, and the \CCNLOS scan radius is $r'$. \textbf{(b)} Each scatterer on the surface of this sphere produces a sinusoidal response with a temporal offset $\dcoffset=r^2 + r'^2$ within a range $[(r-r')^2, (r+r')^2]$. \textbf{(c-1)} The transient sinogram is cropped with this range to recover a 2D image \textbf{(c-2)} via the inverse Radon transform. This is inspired by 2D image recovery with X-ray computed tomography (CT), shown in \textbf{(d-1)} and \textbf{(d-2)}.}
\label{fig:2D_iradon_explain}
\end{figure}

When the hidden object lies on the surface of a sphere of radius $r$ with the same center as the scanning circle (see Fig.~\ref{fig:2D_iradon_explain}(a)), each point on the hidden object produces a sinusoid with the same temporal offset $\dcoffset$, as shown in Fig.~\ref{fig:2D_iradon_explain}(b).  We then recover a 2D image of the object from these measurements by using a standard inverse Radon transform procedure.

We choose a value for the radius $r$, either manually or automatically by computing the mean transient response.  All sinusoids from points on this sphere have a corresponding offset $\gamma = r^2 + r'^2$.  Because the maximum amplitude of sinusoids is $2rr'$, the transient response is contained within a temporal range $\left[\gamma -2rr', \gamma + 2rr'\right] = [(r-r')^2, (r+r')^2]$.  We therefore crop the transient sinogram accordingly, and apply the inverse Radon transform (see Fig.~\ref{fig:2D_iradon_explain}(d-1, d-2))
directly to the results to recover a 2D image (see Fig.~\ref{fig:2D_iradon_explain}(c-1, c-2)).  The pixel coordinate of the recovered image associated for each sinusoid is given by 
\begin{align}
[u,w] &= [\alpha\cos(\beta), \alpha\sin(\beta)] &&\text{(from Radon transform)} \\
&= [2rr'\sin(\theta)\cos(\phi),2rr'\sin(\theta)\sin(\phi)] &&\text{(from Eq.~\eqref{eq:pos_to_param})}\\
&= 2r'[x,y] &&\text{(from Eq.~\eqref{eq:sphericalcoordinates})}
\end{align}
In other words, the recovered image simply represents a scaled orthographic projection of the hidden object onto the relay wall.

\subsubsection{Linear Inversion} 
Our analysis with the Radon Transform demonstrates that significant information about the hidden scene is encoded in a transient sinogram. However, in practice, most common objects are not fully contained within the surface of a sphere, instead consisting of multiple depths and distances from the center of the scanning circle. With that in mind, an important question that arises is whether a \CCNLOS measurement contains enough information to reconstruct these more general objects. To explore this question, we empirically investigate the case of large 2-dimensional planar scenes, which are commonly used by existing NLOS techniques to gauge the accuracy of their reconstructions.

Under this constrained case, efficient recovery of a 2D image $\bm{\rho}_{2D}$ from \CCNLOS measurements involves solving a linear-least squares problem: 
\begin{equation}
    \bm{\rho}_{2D} = \argmin_{\bm{\rho}} \quad \frac{1}{2} || \bm{\tau}_\text{circ} - \mathbf{A}_{d} \bm{\rho} ||_2^2 + \frac{\lambda}{2}\|\bm{\rho}\|^2_2,
\end{equation}
where $\lambda$ controls the weight of the regularization term, and the matrix $\mathbf{A}_d$ represents the mapping from hidden scatterers on a plane $d$ away from the center of the scanning circle, to \CCNLOS measurements.

Both our Radon reconstruction and planar inversion algorithms require knowledge of the sphere or plane containing the hidden object.  If the object is not contained within the surface of the sphere/plane, the recovered images are blurred; we show an analysis in the supplement.  Changing the value for $r$ or $d$ is analogous to a manual refocusing operation that can help produce a clearer image.

\subsection{3D Reconstruction: 3D Imaging via a Modified LCT}
\label{subsec:cmlct}

A natural follow-up question that arises is whether a transient sinogram is sufficient for performing a full 3D reconstruction. Empirically, we show that it is feasible to recover full 3D volumes of the hidden scene from \CCNLOS measurements.  Although this involves solving an underconstrained system due to the limited number of measurements, approximate reconstructions can be achieved by applying non-negativity, sparsity, and total variation priors on the hidden volume, commonly utilized by previous approaches \cite{Ahn2019,Heide2014,Heide2019,o2018confocal}.  

We propose a modified version of the iterative light cone transform (LCT) procedure used in confocal NLOS imaging~\cite{o2018confocal}. Because a \CCNLOS measurement is a subset of a full confocal NLOS measurement, we simply add a sampling term to the iterative LCT procedure, solving the following optimization problem:
\begin{equation} \label{eq:M_in}
    \bm{\rho}_{3D} = \argmin_{\bm{\rho}} \quad \frac{1}{2} \left\Vert \bm{\tau}_{\text{circ}} - \mathbf{MA}\bm{\rho} \right\Vert ^2_2 + \Gamma(\bm{\rho}),
\end{equation}
where the matrix $\mathbf{A}$ maps a 3D volume to a confocal NLOS measurement, the matrix $\mathbf{M}$ subsamples the confocal NLOS measurement to produce a \CCNLOS measurement, and $\Gamma(\cdot)$ represents our non-negativity, sparsity, and total variation priors.  Because the matrix $\mathbf{A}$ can be modelled as a convolution operation, the above expression can be optimized efficiently without having to construct $\mathbf{A}$ explicitly.  We describe our procedure in detail in the supplement.

There are some drawbacks to this formulation. In its current form, it makes no explicit usage of the sinusoidal properties of \CCNLOS measurements that we utilized for object detection and 2D imaging, that could simplify the reconstruction. At the same time, the matrix $\mathbf{M}$ complicates a frequency analysis of the LCT, making it much more unclear which parts of the hidden scene can and cannot be reconstructed. We plan to investigate these phenomena in the future.
\section{Experiments}
\label{sec:experiment}

\noindent
\textbf{Baseline algorithms.}
No existing algorithms operate on \CCNLOS scans, or even just 1D scans.  Thus, we compare our method with two volume reconstruction approaches that rely on full 2D confocal scans: LCT~\cite{o2018confocal} and FK~\cite{Lindell2019}. We identify the peak and compute a maximum intensity projection from each of the output volumes to generate 1D and 2D reconstructions, respectively. To estimate scatterer positions, we also test a trilateration-based approach (``3 Points'') that uses only three scanning points~\cite{o2018confocal}, which we describe in the supplement.

\begin{figure}[tb]
\begin{center}
\includegraphics[width=0.8\textwidth]{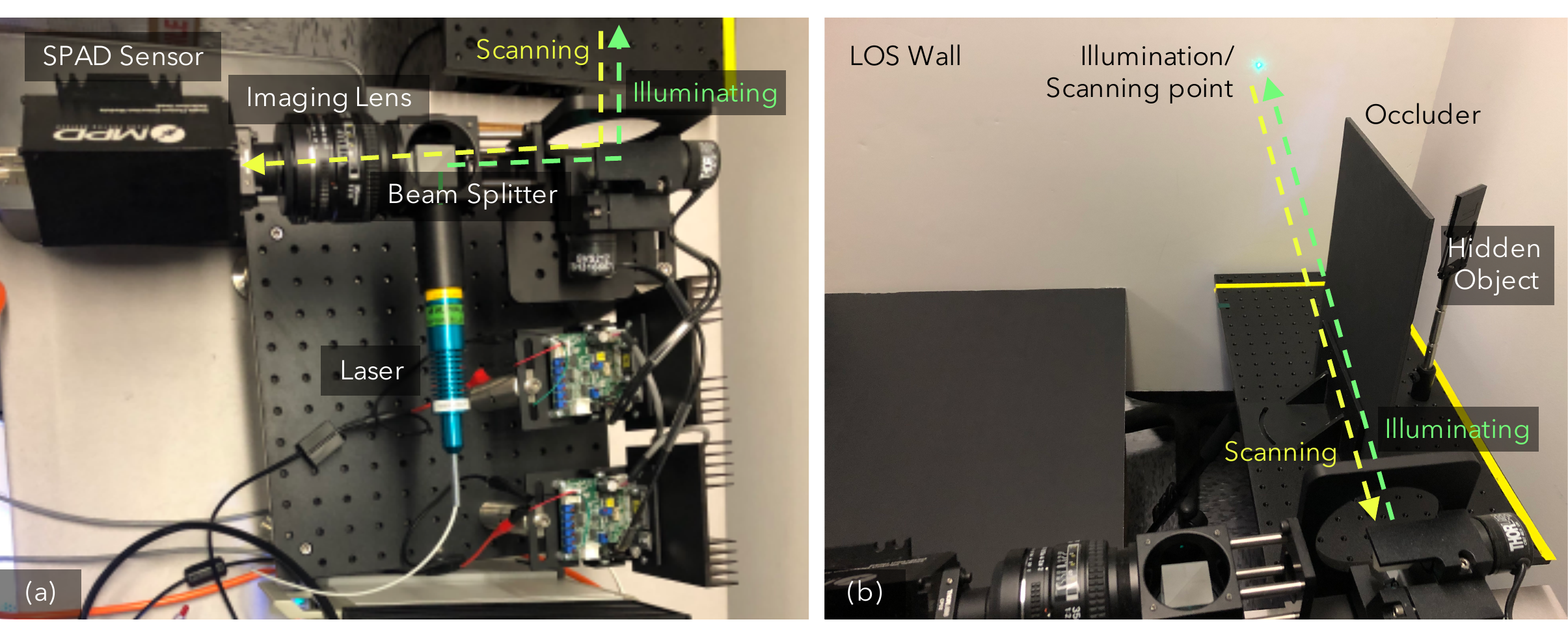}
\end{center}
\caption{\textbf{(a)} Our hardware prototype. \textbf{(b)} A hidden scene with a single NLOS object (a retroreflector) used for our object localization experiments.}
\label{fig:hardware}
\end{figure}

\phantom{}\leavevmode
\noindent
\textbf{Hardware.}
Our prototype \CCNLOS system (Fig.~\ref{fig:hardware}) is based on the system proposed in O’Toole~\etal~\cite{o2018confocal}. Please refer to the supplemental material for more details.
To estimate the computational efficiency of each algorithm, we ran each reconstruction algorithm on a 2017 Macbook Pro  (\SI{2.5}{\giga\hertz} Intel Core i7).

\phantom{}\leavevmode
\noindent
\textbf{Transient Measurement Data.}
For object localization, we use real captured data from our \CCNLOS acquisition system (see Fig.~\ref{fig:1D_tracking_result}(a)), as well as simulated data. In our hardware acquisition system, we captured transients of size $1024(\phi') \times 4096(t)$ from a circular scanning pattern of diameter 1.0m.
To qualitatively evaluate our single-object localization, we synthesized transients of size $64\times64\times2048$ from 100 randomly generated NLOS scenes, where we placed a single scatterer at a random location in a 1.0m$\times$1.0m$\times$1.0m volume 2.0m away from the LOS wall. We used 200 scenes in the two-object case.
For 2D imaging, we used simulated transient data from the Z-NLOS Dataset~\cite{galindo19-NLOSDataset,JaraboSIGA14}, which we resized to $64\times64\times2048$. For 3D imaging, we test our algorithm on real captured data provided by O'Toole~\etal~\cite{o2018confocal}, as well as simulated data from the Z-NLOS Dataset, all rescaled to $64 \times 64 \times 512$. In all cases, to synthesize \CCNLOS data, we sampled 360 angles along the inscribed circle of the confocal grid.

On a typical NLOS hardware setup, the $64 \times 64$ grid data used by FK/LCT would only be captured at roughly 2 Hz. In contrast, \CCNLOS measurements can be captured at 130 Hz, corresponding to just 1.6\% of the capture time.

\subsection{1D Reconstruction: Object Localization}
\label{subsec:result_1d}

Fig.~\ref{fig:1D_tracking_result}(left) demonstrates our methodology for estimating a scatterer's position with real captured transient data, using the approach described in Section~\ref{sec:object_pos_estimation}. Please note that the proposed method is applicable for more than two objects without loss of generality. 

Due to the robustness of Hough voting, our method detects the position of the hidden object(s) even though the transient measurements are quite noisy. See supplemental materials for more results. 

For quantitative validation, Fig.~\ref{fig:1D_tracking_result}(right) compares our method to the baseline approaches using the experimental setup outlined in the previous section. Despite the much smaller number of spatial samples, \CCNLOS was able to achieve similar accuracy to LCT or FK, both of which require an order-of-magnitude more measurements. Computationally, our method was also faster than LCT and FK, but slower than the 3-Points method in the single object case. However, note that the 3-Points method does not generalize beyond a single object.

\begin{table}[t]
    \begin{minipage}{0.6\linewidth}
    	\centering
    	\includegraphics[width=1.0\textwidth]{figure/1D_result.pdf}
    \end{minipage}
    \begin{minipage}{0.36\linewidth}
       \vspace{8pt}
    	\label{tab:trajectory_ablation}
    	\centering
        \resizebox{\textwidth}{!}{
            \begin{tabular}{r@{\hskip 3mm}c@{\hskip 3mm}ccc}
            \toprule
            \multirow{2}{*}{Method}       & Time & \multicolumn{3}{c}{Mean Error} \\
                                          & [sec] & x-axis & y-axis & z-axis       \\
            \midrule 
            LCT~\cite{o2018confocal}      & 6.86                  & 0.84 & 0.89 & 0.19 \\
            FK~\cite{Lindell2019}         & 25.45                 & 0.31 & 0.31 & 0.17 \\
            3 Points~\cite{o2018confocal} & \bf{0.20}             & 0.56 & 0.65 & 0.22 \\
            \CCNLOS (Ours)                & \underline{0.50}      & 0.44 & 0.42 & 0.93 \\
            \bottomrule
            \\
            \toprule
            \multirow{2}{*}{Method}       & Time & \multicolumn{3}{c}{Mean Error} \\
                                          & [sec]  & x-axis & y-axis & z-axis       \\
            \midrule 
            LCT~\cite{o2018confocal}      & 6.92                  & 1.34 & 1.47 & 1.40 \\
            FK~\cite{Lindell2019}         & 26.65                 & 0.29 & 0.39 & 0.07 \\
            3 Points~\cite{o2018confocal} & ---                   & ---  & ---  & ---  \\
            \CCNLOS (Ours)                & \bf{0.52}             & 2.20 & 1.30 & 7.37 \\
            \bottomrule     
            \end{tabular}}
    \end{minipage}
	
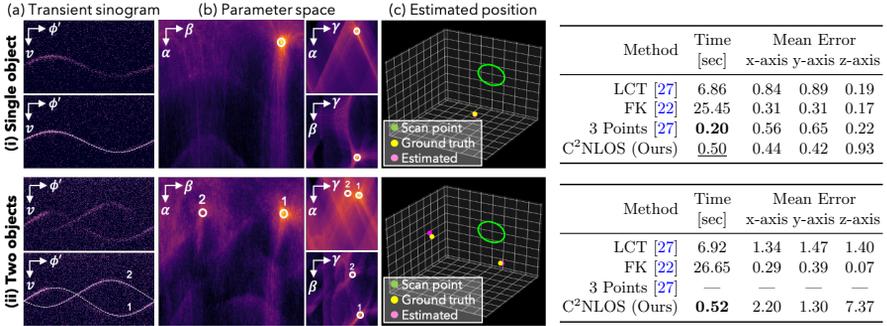
\captionof{figure}{Estimating the position of (i) one and (ii) two scatterers. 
	\textbf{Left:} Given a transient measurement in (a) top, we generate a sinusoid parameter space (b). The sinusoid parameters that best fit the transient sinogram (see (a) bottom) are obtained by finding its peak~(see annotations on (b)). The 3D position of the object is reconstructed with the estimated parameters. 
    \textbf{Right:} Quantitative evaluation with mean estimation error and computational time. Despite only requiring roughly 1.6\% of the capture time of LCT or FK, \CCNLOS estimated the position within almost the same order of accuracy as the other methods. Our approach was also faster than LCT and FK.}
    \label{fig:1D_tracking_result}
\end{table}

\begin{figure}[t!]
\centering
\begin{minipage}[c]{1\textwidth}
\centering
\includegraphics[width=0.85\textwidth]{figure/2D_result.pdf}
\caption{Quantitative and qualitative results on 2D imaging on large planar scenes. Despite sampling far fewer measurements than LCT and FK, both of our inverse Radon reconstruction/linear inversion-based methods reconstructed images that were similar in quality. The SSIM scores for the proposed methods were slightly worse than LCT or FK. However, our methods were much more computationally efficient (\eg, inverse Radon reconstruction yielded a 50x speedup over LCT).}
\label{fig:2d_result}
\end{minipage}
\begin{minipage}[c]{1\textwidth}
\centering
\includegraphics[width=1.0\textwidth]{figure/radon2d_diff_focal.pdf}

\caption{Our inverse Radon reconstruction-based 2D imaging with different focus planes (right focuses towards larger depth). Even for non-planar objects like a bunny, \CCNLOS measurements contain sufficient information about the hidden object for an inverse Radon transform to approximately reconstruct its visual appearance.}
\label{fig:2d_focalplane}
\end{minipage}
\end{figure}

\subsection{2D Reconstruction: 2D Plane Imaging}
\label{subsec:result_2d}

Fig.~\ref{fig:2d_result} shows the qualitative and quantitative 2D imaging results on large planar scenes. Despite requiring just 1.6\% of the capture time, the linear inversion method was able to visualize the hidden image plane. At the same time, even though our inverse Radon reconstruction approach was not designed for large planar objects, it still recovers an approximate reconstruction of the 2D scene. 

As we mentioned in Section~\ref{sec:inverse_radon_method}, changing the value for the radius $r$ of the sphere or distance $d$ is analogous to a manual refocusing operation. Fig.~\ref{fig:2d_focalplane} shows the results with the proposed Radon reconstruction-based method, in which the results with larger $r$ (farther from the wall) are shown to the right. Our method was able to visualize not only flat objects, but also objects with a wider range of depths, as shown by the \textit{bunny} scene.

For quantitative validation, we used SSIM~\cite{SSIM} as a metric. As shown in Fig.~\ref{fig:2d_result}, both of our 2D imaging methods yield slightly worse results compared to LCT and FK. However, both our Radon-reconstruction and linear-inversion procedures are significantly faster than LCT and FK.

\begin{figure}[t]
\begin{center}
\includegraphics[width=1.0\textwidth]{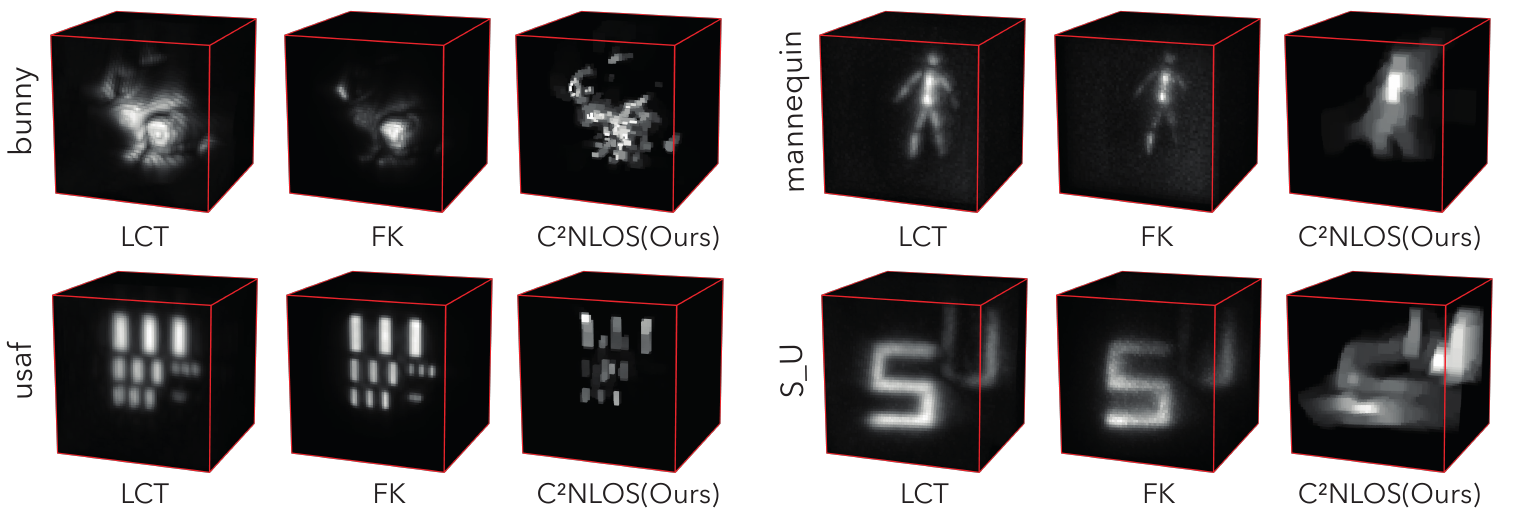}
\end{center}
\caption{3D volume reconstruction results. Even though a \CCNLOS scan requires just 1.6\% of the capture time needed by LCT and FK, our approach still generates an approximate reconstruction. More results in the supplement.}
\label{fig:3D_result}
\end{figure}

\subsection{3D Reconstruction: 3D Volume Imaging}
\label{subsec:result_3d}
We use the Alternating Direction Method of Multipliers (ADMM) \cite{boyd2011} to minimize the optimization problem from Equation~\eqref{eq:M_in}. We show a full derivation in the supplement. Because of the iterative optimization process required by our strong priors, our reconstruction operator is inherently slower than both FK and LCT.  However, our method yields similar runtimes to the iterative versions of LCT and the Gram operator \cite{Ahn2019}, both of which use a similar optimization formulation.

In order to evaluate our 3D reconstructions, we test LCT, FK, and our modified LCT procedure on a variety of different scenes in Fig.~\ref{fig:3D_result}. In general, while FK and LCT demonstrate much higher reconstruction quality, our approach still captures important features of the hidden scene, like the presence and depth of multiple planar objects in the \textit{S\_U} scene and the overall pose in the \textit{mannequin} scene. Empirically, this shows that significant volumetric information of the hidden scene can be recovered from a single transient sinogram.

\section{Conclusion}

We show that a transient sinogram acquired through \CCNLOS scanning is sufficient for solving a number of imaging tasks, even though the dimensionality of the measurement is smaller than those captured by existing NLOS methods. Through an analysis of the image formation model, we explain how the measurements are fundamentally sinusoidal and lend themselves to efficient reconstruction algorithms, including a Hough voting procedure for estimating the 3D position of scatterers and an inverse Radon technique for recovering 2D images of hidden objects.  We empirically demonstrate that the measurements can also be applied to recover full 3D volumes.  We believe these contributions mark a significant step in our understanding of \emph{efficient} imaging techniques for revealing objects hidden just around a corner.

\noindent \\
\textbf{Acknowledgements.}
We thank Ioannis Gkioulekas for helpful discussions and feedback on this work. M. Isogawa is supported by NTT Corporation.  M. O'Toole is supported by the DARPA REVEAL program.

\newpage

\bibliographystyle{splncs04}
\bibliography{egbib}

\onecolumn

\newpage
\newpage

\appendix
\noindent
\large
\textbf{Appendix}
\normalsize

\section{Prototype Confocal NLOS Imaging System}

\begin{figure}[h]
\begin{center}
\includegraphics[width=0.8\textwidth]{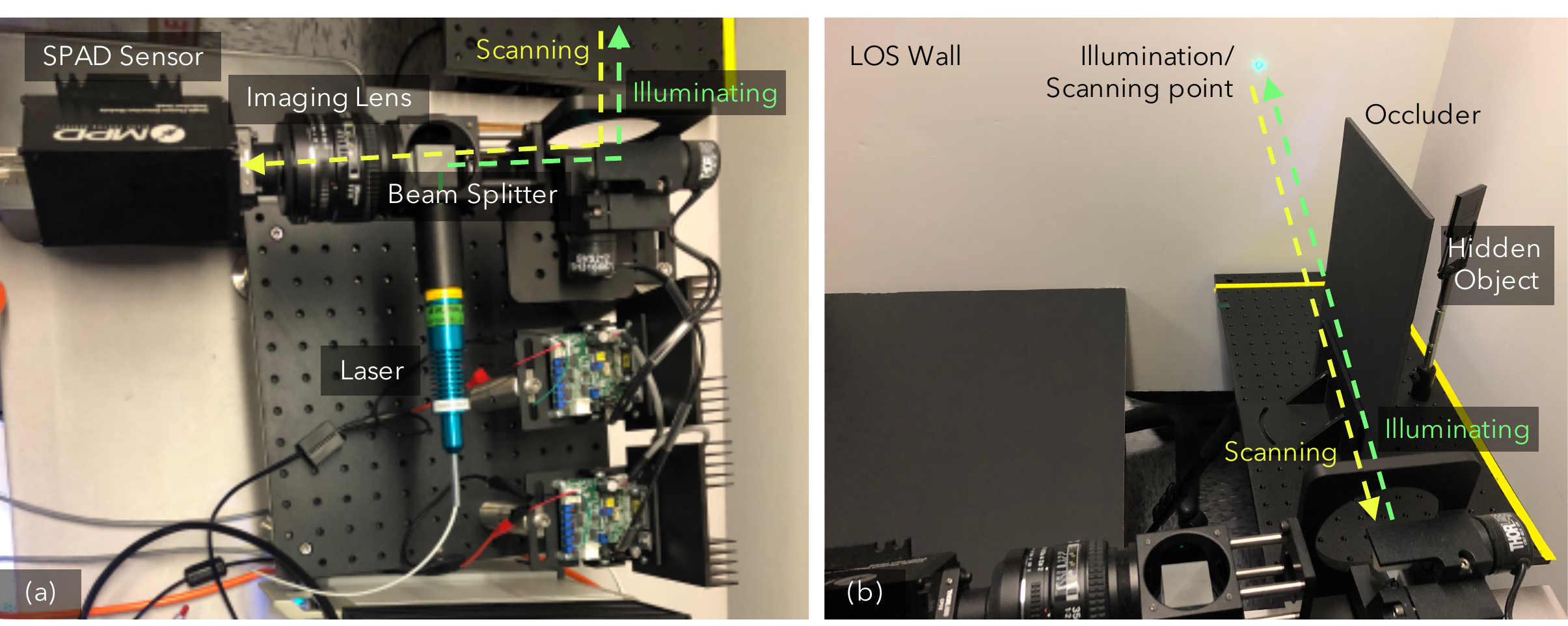}
\end{center}
\vspace{-10pt}
\caption{\textbf{(a)} Our hardware prototype. \textbf{(b)} A hidden scene with a single NLOS object (a retroreflector) used for our object localization experiments.}
\vspace{-2mm}
\label{fig:hardware1}
\end{figure}

The design of our prototype confocal NLOS imaging system (shown in Figure~\ref{fig:hardware1}) is based on  the system proposed in O'Toole~\etal~\cite{o2018confocal}.  Our laser is a low-power picosecond pulse diode from ALPHALAS with a wavelength of \SI{520}{\nano\meter}, a full width at half maximum (FWHM) of \SI{60}{\pico\second}, and a peak power of \SI{280}{\milli\watt}.  The laser emits pulses of light at a rate of \SI{10}{\mega\hertz}.  A fast-gated single photon avalanche diode (SPAD) from Micro Photon Devices (MPD) measures the response; the gate feature of the SPAD is turned off in our experiments.  Our time-correlated single photon counting system (TCSPC) is a PicoHarp 300 from PicoQuant, and its role is to convert the SPAD's output into a stream of photon events.  A MATLAB script then bins these photon events into a transient sinogram.  The laser and SPAD are aligned with a beamsplitter (Thorlabs PBS251), and a Nikon lens focuses the light from the scene onto the SPAD.

Another MATLAB script interfaces with a National Instruments Data Acquisition Device (NI-DAQ USB-6343) to control a pair of large beam galvo mirrors (Thorlabs GVS012).  The mirrors control the point on a wall illuminated with laser light, and measured by the SPAD.  After calibrating for the position of the wall relative to our setup, the galvo mirrors continuously scan a circle on the wall of a user-specified radius at a rate of \SI{130}{\hertz}.

\vspace{-5mm}

\section{Additional Analysis and Results}
\label{sec:results}

\subsection{1D Reconstruction: Additional Object Localization Results with Real Captured Data}
\label{sec:1d_real_results}

\begin{figure}[h]
\begin{center}
\includegraphics[width=1\textwidth]{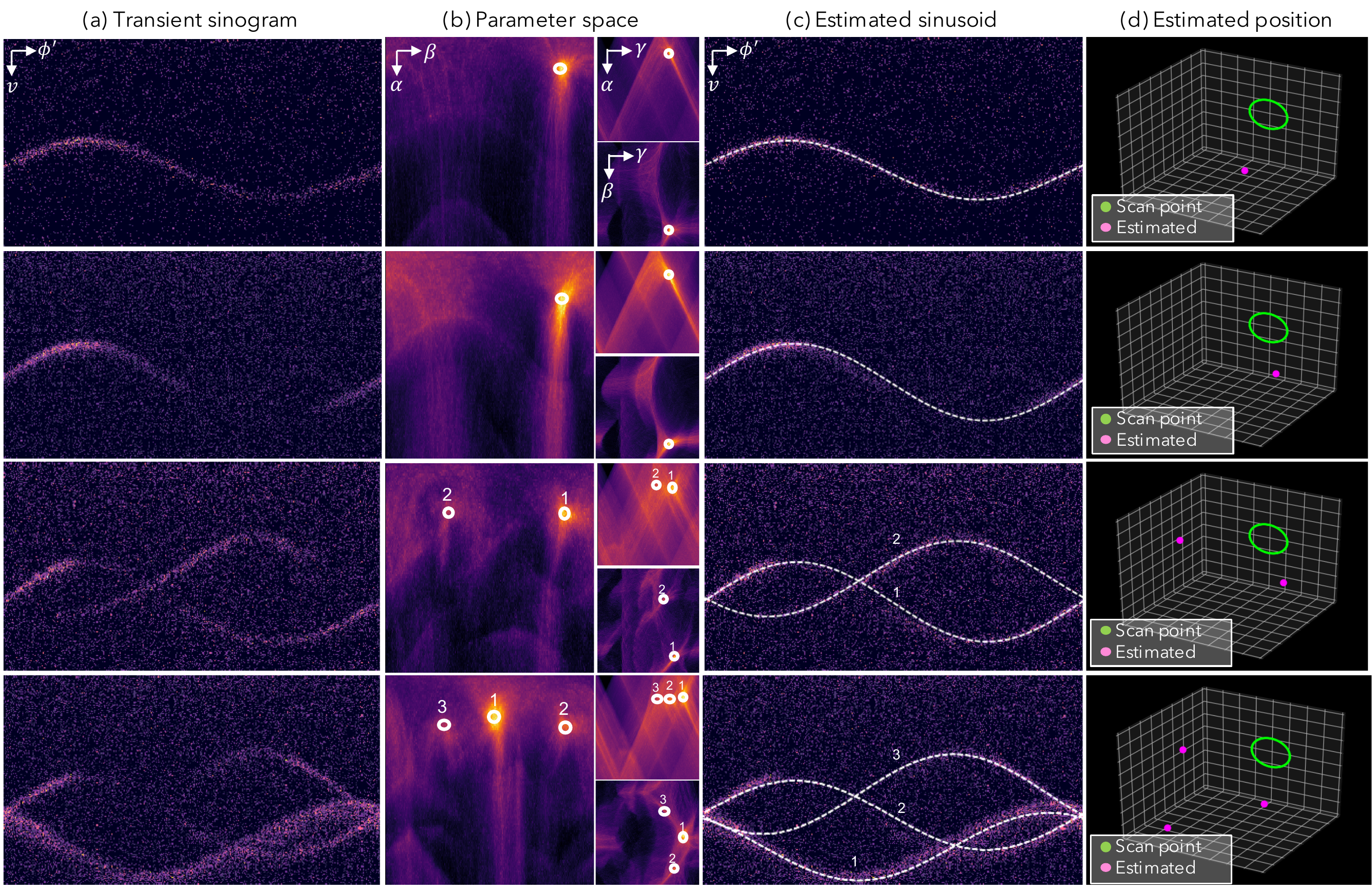}
\end{center}
\vspace{-10pt}
\caption{Additional experimental results for estimating the position of scatterers. \textbf{(a)} Transient sinograms of different numbers of scatterers.  \textbf{(b)} From every transient sinogram, we estimate the amplitude, phase, and offset of the sinusoids using a Hough transform procedure.  \textbf{(c)} By identifying peaks in the parameter space shown in (b), our approach recovers the sinusoid parameters that best fit the measured transient sinograms. \textbf{(d)} The 3D position of the object is finally reconstructed with the estimated parameters. }
\vspace{-2mm}
\label{fig:1d_result}
\end{figure}

Figure~\ref{fig:1d_result} expands upon the object localization results shown in Figure~6 of the main paper.  Here, we demonstrate the ability to estimate the positions of one, two, or three scatterers hidden from direct line-of-sight with our prototype NLOS system.  Even in the case of three scatterers, our Hough voting approach accurately estimates the parameters (amplitude $\alpha$, phase $\beta$, and offset $\gamma$) of the three corresponding sinusoids from the transient sinogram.  We then convert the recovered parameters to each object's 3D position.

\subsection{2D Reconstruction}
\label{sec:2d_results}

\subsubsection{The Radon Transform and 2D NLOS Imaging}
\label{sec:2d_radon_analysis}
\begin{figure}[h]
\begin{center}
\includegraphics[width=1\textwidth]{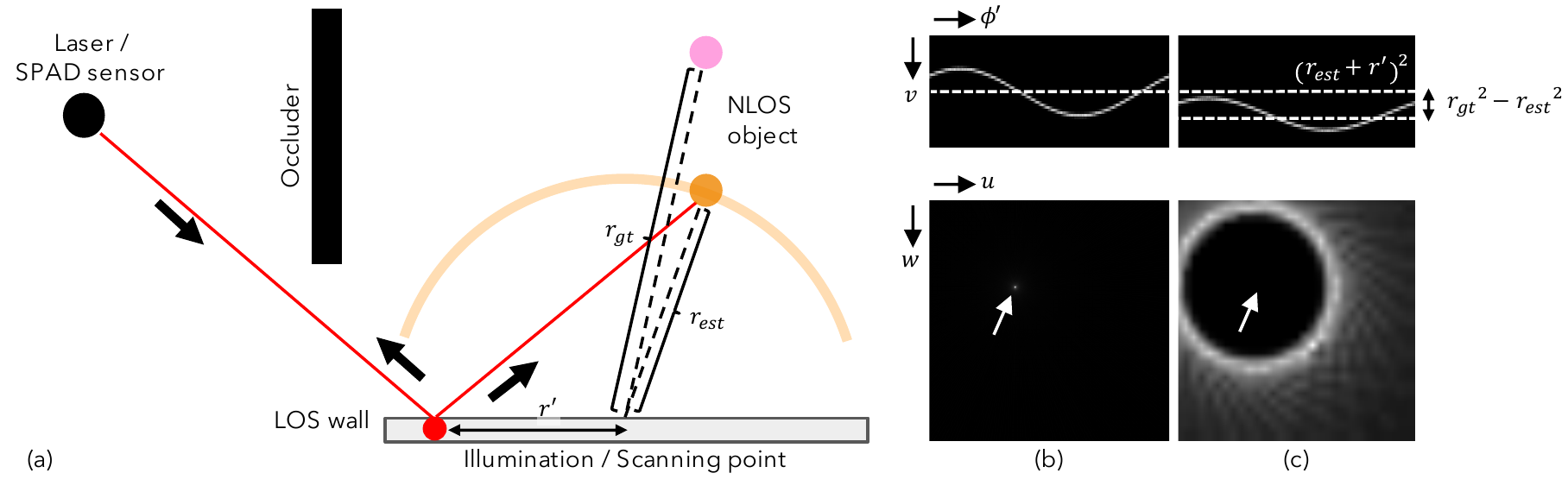}
\end{center}
\caption{2D imaging via the inverse Radon transform. \textbf{(a)} An infinitesimally small NLOS scatterer positioned at $[x, y, z]$ is a distance $r_\text{gt}$ away from the center of the scanning circle with radius $r'$. \textbf{(b)} If the estimated distance $r_\text{est}$ perfectly matches $r_\text{gt}$ (\ie, $r_\text{est} = r_\text{gt}$), the reconstruction contains a single point $2r'[x, y]$ representing the position of the scatterer.  \textbf{(c)} Suppose the distance is incorrectly estimated (\ie, $r_\text{est} \neq r_\text{gt}$). Then, the sinusoid corresponding to the scatterer will be shifted by $r_\text{gt}^2 - r_\text{est}^2$ in the transient sinogram.  After applying the inverse Radon transform, the reconstruction is a circle centered at $2r'[x, y]$ with a radius proportional to the error in distance.  We refer to this circle as the \textit{Radon circle of confusion}.}
\label{fig:iradon_proof}
\end{figure}

Let's consider the case of a single scatterer, located at some distance $r_\text{gt}$ away from the center of the scanning circle (see Figure~\ref{fig:iradon_proof}(a)). As shown in Figure~\ref{fig:iradon_proof}(b), this scatterer will contribute a sinusoid to the transient measurements with the following form:

\begin{eqnarray}
v(\phi') = r_\text{gt}^2 + r'^2 - 2 r_\text{gt}r'\sin(\theta)\cos(\phi-\phi') = \dcoffset - \amplitude \cos(\phase - \phi'),
\end{eqnarray}
where
\begin{eqnarray}
    \amplitude = 2 r_\text{gt}r'\sin(\theta), \quad\phase = \phi, \quad\dcoffset =  r_\text{gt}^2 + r'^2.
\end{eqnarray}

Consider the case where we assume the correct distance is $r_\text{est}$, which may or may not equal $r_\text{gt}$. Following the methodology outlined in the paper, we recenter our transient sinogram at offset $r_\text{est}^2 + r'^2$. For simplicity, we ignore the effect of cropping for the following theoretical analysis. This formulation has a number of implications, which we describe below.

Now consider the case where $r_\text{est} = r_\text{gt}$. Then, the sinusoid of the scatterer is perfectly centered as input to the inverse Radon reconstruction. Because a single spatial point maps exactly to a perfectly centered sinusoid under the Radon Transform \cite{kak2002principles}, our output 2D image captures a perfect scaled orthographic projection of the scatterer, located at coordinate $2r'[x,y]$ as given in Equation~(13) in the main paper.

Now, suppose $r_\text{est} \neq r_\text{gt}$. The sinusoid of the scatterer will be instead shifted to be centered at $\Delta = r_\text{gt}^2 - r_\text{est}^2$  (see Figure~\ref{fig:iradon_proof}(c)). Let $G_\beta(\omega)$ be the 1D Fourier Transform of the resampled transient measurement for a spatial circular sample $\beta$. Because every transient measurement is shifted by the same amount $\Delta$, $G_\beta(\omega) = e^{-i \omega \Delta} F_\beta(\omega)$, where $F_\beta(\omega)$ is the 1D Fourier Transform of a correctly shifted measurement and $e^{-i \omega \Delta}$ is the same complex exponential for every $\beta$. 
By the projection-slice theorem, $G_\beta(u)$ exactly gives the 1D slice with angle $\beta$ through the origin of the 2D Fourier Transform of the output image \cite{kak2002principles}. Therefore, the 2D Fourier Transform of this incorrect scene $G(u, v)$ will equal the 2D Fourier Transform of a correctly-shifted scene $F(u, v)$, modulated by the sinusoid given by $e^{-i \omega \Delta}$ rotated about the $\mathbf{0}$-frequency origin. In the spatial domain, this circular sinusoidal pattern corresponds to a convolution with a circular kernel with radius $\Delta$. In effect, rather than corresponding to a single point in the output 2D image from an inverse Radon Transform, the scatterer instead maps to a circle with radius $\Delta$ centered at $2r'[x,y]$. The radius $\Delta$ of this \textit{Radon circle of confusion} does not change with the scanning circle radius $r'$.

Under this analysis, the 2D Radon NLOS image formation model for estimated distance $r_\text{est}$ in the Fourier domain is given by

\begin{equation}
    O_{r_\text{est}}(u, v) = 2r'\sum_{d} e^{-i \sqrt{u^2 + v^2} (r_d^2 - r_\text{est}^2)} F_d(2r'u, 2r'v) ,
\end{equation}
where $F_d(u, v)$ is the 2D Fourier Transform of the orthographic projection onto the wall of all elements of the scene $r_d$ away from the center of the scanning circle. The $2r'$ term accounts for the scaling given by Equation~(13).

This model has the following implications:

\begin{enumerate}
    \item An object that perfectly lies upon the surface of a sphere with known radius yields a perfect reconstruction via an inverse Radon Transform, because every object point will be perfectly orthographically projected in the output image.
    \item In the more general case, object points $r_d$ away that do not satisfy the sphere constraint will generate circular patterns with radius $r_d^2 - r_\text{est}^2$ centered at their scaled orthographic projections.
\end{enumerate}

We explore these effects in further detail in the rest of this section. In Section~\ref{sec:2d_w_sphere_constraint}, we demonstrate the effectiveness of inverse Radon reconstruction for scenes that perfectly lie upon the surface of a sphere. In practice, because many common scenes do not satisfy this spherical constraint, we propose either empirical undistortion (Section~\ref{sec:undistortion}) or larger scanning circles (Section~\ref{sec:circle_size}) to remedy the artifacts that arise from a direct application of inverse Radon reconstruction. We also show manual refocusing via changing the estimated radius $r_\text{est}$ in Section~\ref{sec:manual_focus}.

To implement our inverse Radon reconstruction, we use MATLAB's \textit{iradon} functionality, which uses a backprojection operation. We discuss the empirical effects of filtered backprojection in 2D NLOS imaging in Section~\ref{sec:filter}.

\subsubsection{2D Image Reconstruction of Spherically-Constrained Scenes}
\label{sec:2d_w_sphere_constraint}

Section~5.2 in the main paper shows that even though our inverse Radon reconstruction approach was not designed for large planar objects, it still recovers an approximate reconstruction of the 2D scene. 
With the above in mind, one might be interested in the quality of reconstruction when the \textbf{scene actually satisfies the sphere constraint} (\ie, when the hidden object lies on the surface of a sphere of radius $r$ with the same center as the scanning circle).

\begin{figure}[h]
\begin{center}
\includegraphics[width=1.0\textwidth]{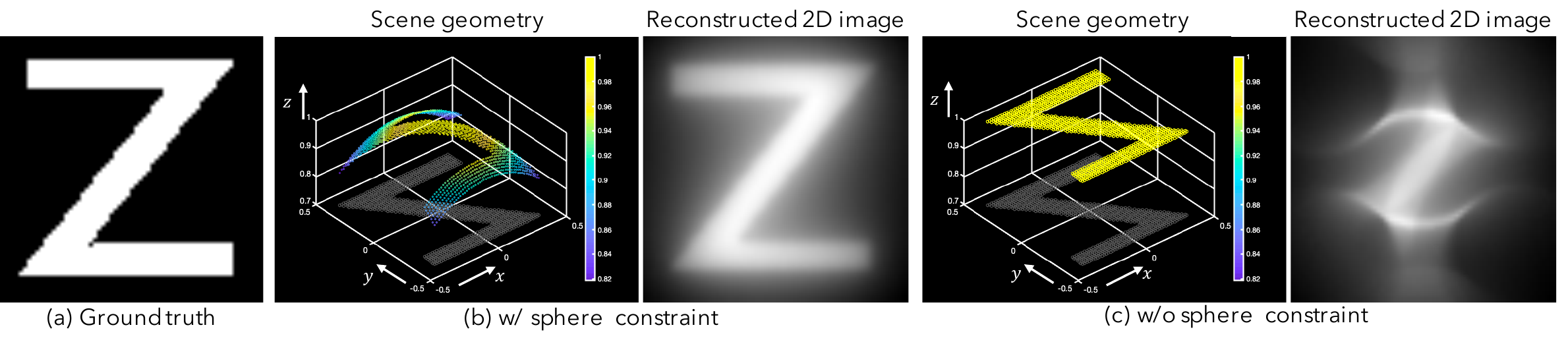}
\end{center}
\vspace{-10pt}
\caption{2D imaging spherically-constrained scenes. \textbf{(a)} Ground truth projection. \textbf{(b)} If every point of the hidden scene lies within the surface of a sphere, our method reconstructs an accurate 2D image. \textbf{(c)} Even though our inverse Radon reconstruction approach was not designed for large planar objects that violate the sphere constraint, it still recovers an approximate reconstruction of the 2D scene.}
\vspace{-2mm}
\label{fig:2D_shpere_constraint}
\end{figure}

Figure~\ref{fig:2D_shpere_constraint} shows 2D reconstruction results for a simulated Z shape \SI{1.0}{\meter} from the wall, in which the sphere constraint is fully satisfied (Figure~\ref{fig:2D_shpere_constraint}(b)) or not satisfied (Figure~\ref{fig:2D_shpere_constraint}(c)). With a \SI{1.0}{\meter} scanning diameter, the spherically-constrained scene is much more accurately reconstructed compared to the planar version, as expected.

\subsubsection{2D Image Undistortion when the Scene is not Spherically-Constrained}
\label{sec:undistortion}

\begin{figure}[h]
\begin{center}
\includegraphics[width=0.8\textwidth]{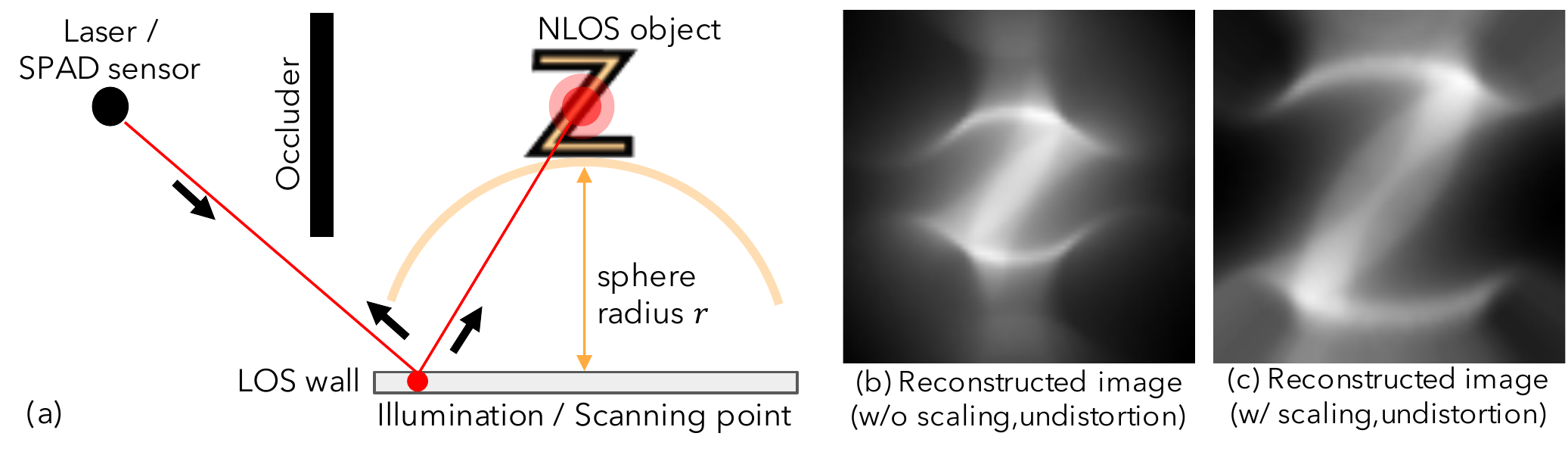}
\end{center}
\vspace{-10pt}
\caption{Undistorting 2D images of a scene that does not satisfy the sphere constraint. \textbf{(a)} Illustration of \CCNLOS scene geometry.  \textbf{(b)} When the object does not perfectly lie on the surface of a sphere, the output of an inverse Radon reconstruction suffers from circular artifacts. \textbf{(c)} Empirically, we observe that applying a fisheye lens undistortion procedure improves the quality of the reconstruction.}
\vspace{-2mm}
\label{fig:iradon_undistortion}
\end{figure}

As shown in Figure~\ref{fig:iradon_undistortion}(a), when the NLOS object does not satisfy the sphere constraint, the inverse Radon transform produces a distorted version of the Z object. For every point on the Z object, the inverse Radon transform produces a circle with a radius proportional to the distance of that point from the surface of the sphere.  This results in large circular distortions when a planar object is too large relative to the size of the sphere, as shown in Figure~\ref{fig:iradon_undistortion}(b). To compensate for this distortion, we empirically found that a simple fisheye lens undistortion operation and a cropping operation helps to produce a clearer image in such cases.

\subsubsection{Effect of the Scanning Circle's Radius}
\label{sec:circle_size}

As shown in Section \ref{sec:2d_radon_analysis}, the size of the \textit{Radon circle of confusion} does not depend on the size of the scanning circle. However, because the rest of the image scales linearly with scanning radius $r'$ from Equation~(13), the effective size of the Radon circle of confusion in relation to the other features of the hidden scene therefore decreases with $r'$. Therefore, to minimize the effects of an incorrect $r_\text{est}$, we should maximize $r'$.

\begin{figure}[h]
\begin{center}
\includegraphics[width=1.0\textwidth]{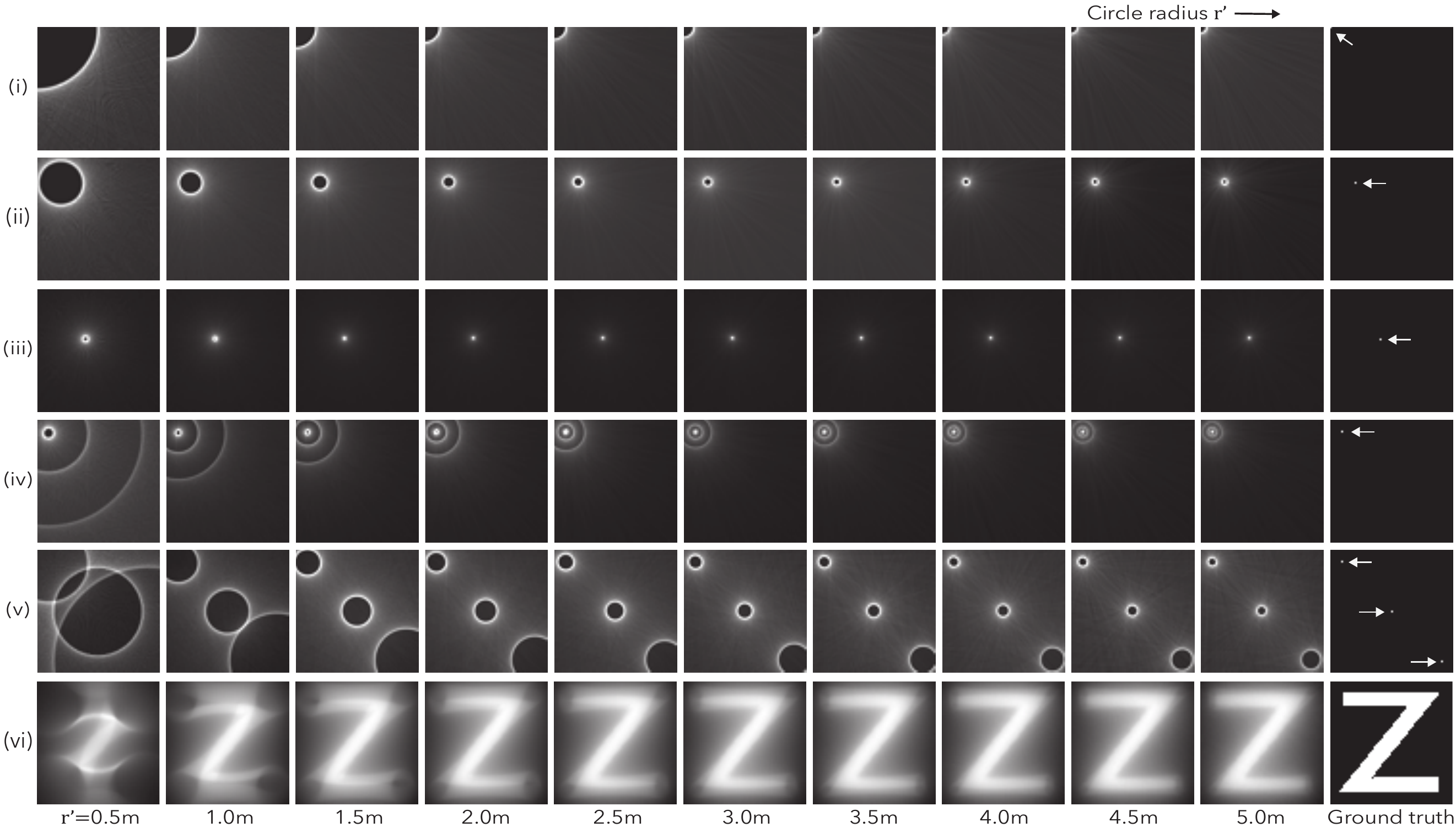}
\end{center}
\vspace{-10pt}
\caption{2D NLOS imaging with different scanning circle sizes (right shows results with larger scanning circles). The effective size of the Radon circle of confusion decreases with larger scanning circle radius $r'$, resulting in higher image quality.}
\label{fig:circle_size}
\end{figure}

Figure~\ref{fig:circle_size} shows reconstructed 2D images with different circle scanning sizes (right shows results with larger scanning circles). Figures~\ref{fig:circle_size}(i) to (iii) show reconstructed 2D images for a scene with a single infinitesimally small scatterer at $(x,y,z)=(0.5, 0.5, 1.0) [m]$, $(0.3, 0.3, 1.0) [m]$, and $(0.1, 0.1, 0.1) [m]$, respectively. 
Figure~\ref{fig:circle_size}(iv) and (v) contain three scatterers at different depths. In Figure~\ref{fig:circle_size}(iv), the scatterers share the same $x,y$ location $(0.4, 0.4)$ but are positioned at different depths $z = 0.8$, $0.4$, and $1.2$. In Figure~\ref{fig:circle_size}(iv), the scatterers are positioned at $(x,y,z)=(0.4, 0.4, 1.0)$, $(0.0, 0.0, 0.8)$, and $(-0.4, -0.4, 1.2)$. As shown in these figures, scatterers at different depths produce circles of confusion of different sizes, but centered at the scatterers' $x,y$ locations. Using a larger scanning radius $r'$ reduces the relative sizes of the circles of confusion.%

Figure~\ref{fig:circle_size}(vi) shows reconstruction results for a simulated Z shape \SI{1.0}{\meter} from the wall. \textbf{The larger the scanning circle, the smaller the effect of the circles of confusion, resulting in clearer images.} In practice, light falloff severely reduces the quality of the signal as $r'$ increases. Thus, a \CCNLOS imaging system should aim to find the right balance between SNR and scanning circle size.

\subsubsection{Synthetic Refocusing}
\label{sec:manual_focus}

\begin{figure}[h]
\begin{center}
\includegraphics[width=1.0\textwidth]{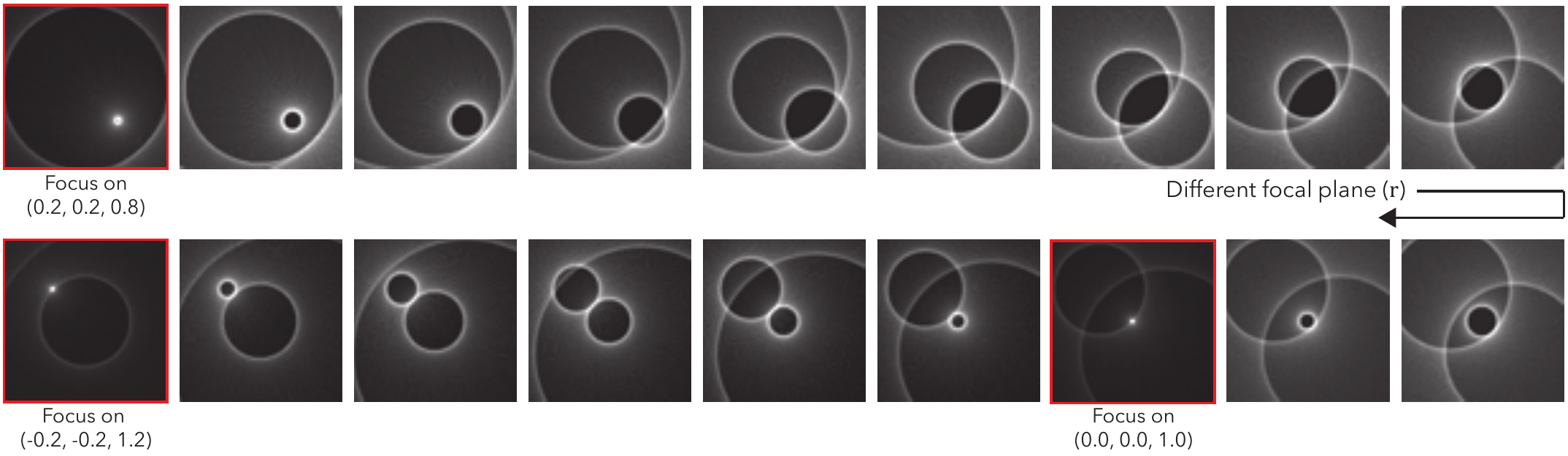}
\end{center}
\vspace{-10pt}
\caption{Manual refocusing in 2D NLOS imaging on a 3-scatterer scene. By changing the estimated radius $r$, we can manually refocus to one of the scatterers, while making the circles of confusion for the other scatterers larger.}
\label{fig:manual_refocus}
\end{figure}

As mentioned in Section~5.2 of the main paper, changing the value for the estimated sphere radius $r$ can be used to focus on different parts of the hidden scene. Figure~\ref{fig:manual_refocus} shows refocusing with three scatterers at different depths, located at $(x, y, z)=(0.2, 0.2, 0.8)$, $(0.0, 0.0, 1.0)$, and $(-0.2, -0.2, 1.2)$. As shown in the figure, we can adjust the value of $r$ to \emph{``focus''} the image at a particular radius, which \emph{``blurs''} points at other radii with a larger circle of confusion.

\vspace{-12mm}

\begin{figure}[h]
\begin{center}
\includegraphics[width=0.8\textwidth]{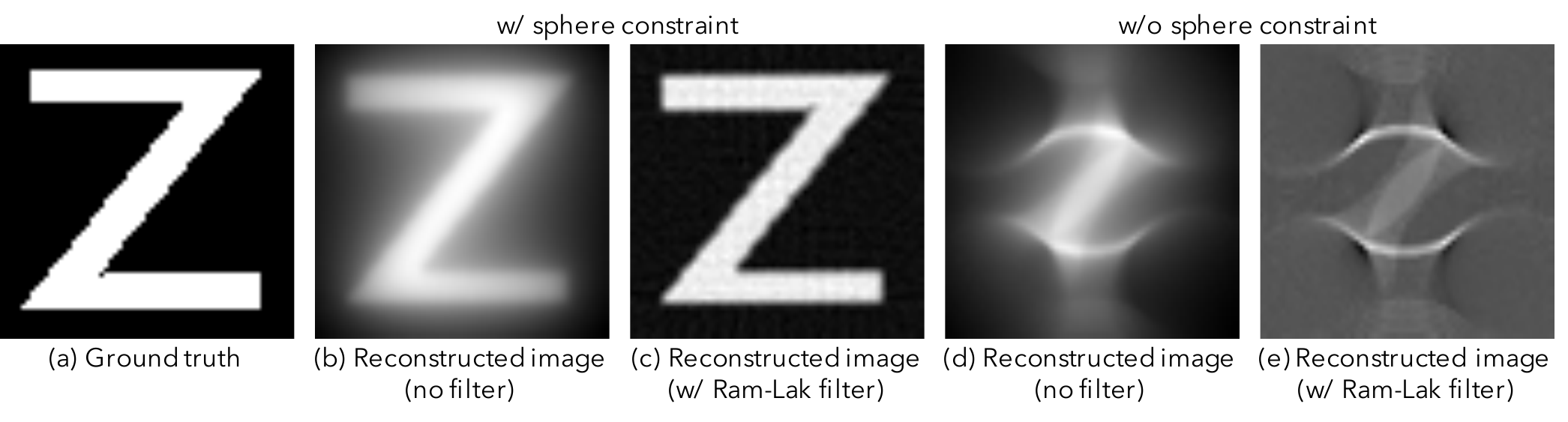}
\end{center}
\vspace{-10pt}
\caption{While traditional inverse Radon reconstructions rely on high pass filters, we avoid them in 2D NLOS imaging because such filters empirically enhance the effects of the Radon circle of confusion, making the artifacting much more visible. When the hidden object satisfies the sphere constraint, a Ram-Lak filter generates a high quality reconstruction as shown in \textbf{(c)}. However, when the object does not perfectly satisfy the spherical constraint as in \textbf{(d)} and \textbf{(e)}, we find that the non-filtered version empirically produces nicer results.}
\vspace{-2mm}
\label{fig:filter}
\end{figure}

\subsubsection{Frequency-Domain Filtering for Inverse Radon Reconstruction}
\label{sec:filter}
 An inverse Radon transform procedure often uses frequency-domain filtering to attenuate low-frequency components that are over-represented in the measurements.  In the absence of noise with an object that perfectly satisfies the spherical constraint, a ramp filter (also known as a Ram-Lak filter) can perfectly reconstruct the hidden scene~\cite{kak2002principles}; see Figure~\ref{fig:filter}(c).  However, as shown in Figure~\ref{fig:filter}(e), we empirically find that these filters typically enhance the high-frequency circular artifacts generated by objects that do not satisfy the spherical constraint. As a result, we choose to use an unfiltered version of the inverse Radon transform for our 2D reconstructions, as demonstrated in Figure~\ref{fig:filter}(d).

\subsubsection{Relationship between reconstruction quality and number of samples compared with 2D grid scanning}

To investigate the impact on number of samples for the Radon reconstruction when objects satisfy the sphere constraint, we refer to~\cite{kak2002principles} (end of Section 5.1.1), which states that ``for a well-balanced N x N reconstructed image, […] the total number of projections should also be roughly N'', corresponding to $N$ samples on the scanning circle. \CCNLOS requires far fewer samples when compared to the $N \times N$ grid required by LCT~\cite{o2018confocal} and FK~\cite{Lindell2019}. This also implies that more circle samples allow for higher resolution reconstructions; however, too many samples yield diminishing returns, because the transients have limited temporal resolution. 

\begin{figure}[h]
\begin{center}
\includegraphics[width=0.6\textwidth]{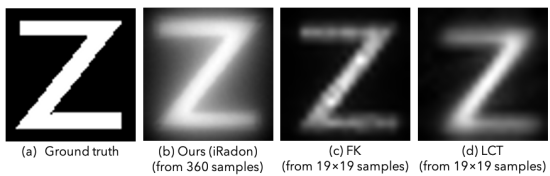}
\end{center}
\vspace{-10pt}
\caption{Qualitative comparison of 2D reconstruction quality for a fixed budget of 360 samples. For FK/LCT, we perform reconstructions from a $19 \times 19$ grid of samples (361 total samples). In this scenario, while both scanning patterns use the same number of samples, a \CCNLOS pattern can be acquired 19 times faster.}
\vspace{-2mm}
\label{fig:19x19reconstruction}
\end{figure}

As we describe in the main paper, our \CCNLOS data consist of 360 samples and are used to reconstruct 2D images at a resolution of $360 \times 360$. However, for a budget of 360 samples, LCT and FK would be limited to $19 \times 19$ spatial samples, resulting in only a $19 \times 19$ reconstruction. For a fixed budget of 360 samples, we show a qualitative comparison in Figure~\ref{fig:19x19reconstruction} for 2D reconstruction and Figure~\ref{fig:3d_19} for 3D reconstruction. In both 2D and 3D, we believe \CCNLOS is comparable in quality to LCT and FK.

However, it is important to note that the number of samples is not the bottleneck. Rather, acquisition speeds are fundamentally limited by the scanning path, and how quickly the mirror galvanometers can follow this path. A single row of a coarse grid requires the same capture time as an entire circular scan; therefore, sampling a $19 \times 19$ grid is still $19$ times slower than a \CCNLOS scan.

\section{Reconstruction Procedure Details}
\label{sec:procedures}
This section explains various implementation details for the algorithms described in the paper. Section ~\ref{sec:hough_vote} explains Hough voting for object localization in further detail. Section~\ref{sec:3points} describes the 3 Points algorithm~\cite{o2018confocal}, a baseline for single object localization used in Section~5.1. In Section~\ref{sec:admm}, we show a full derivation of the proximal operators used for our 3D volume imaging described in Section~5.3 in the main submission.

\subsection{Sinusoid Parameter Estimation based on the Hough Transform}
\label{sec:hough_vote}

As mentioned in Section~4.1 of the main paper, our method estimates the best matching amplitude $\amplitude$, phase $\phase$, and offset $\dcoffset$ for every sinusoid in a transient sinogram. This section aims to provide more detail for this procedure. 

\begin{figure}[h]
\begin{center}
\includegraphics[width=1.0\textwidth]{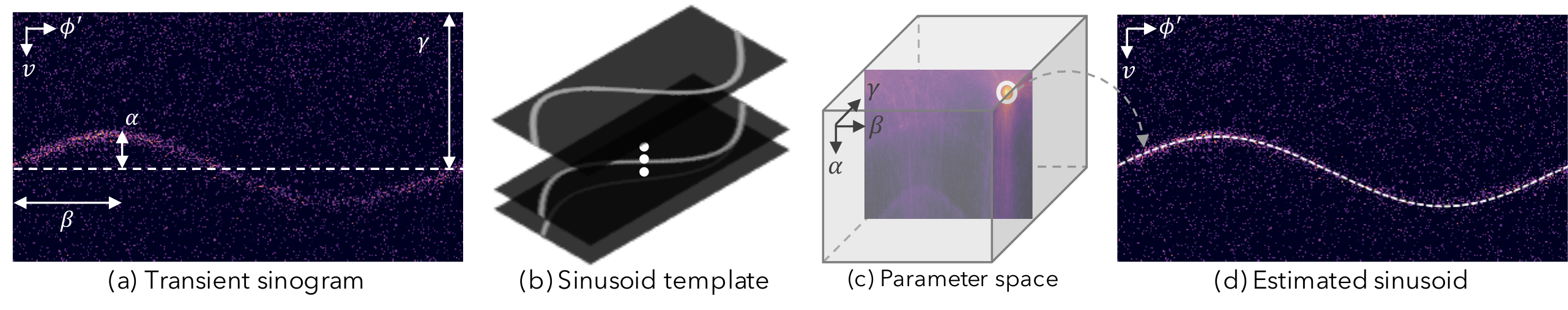}
\end{center}
\vspace{-10pt}
\caption{Our Hough transform-based object localization. Given a transient sinogram \textbf{(a)}, we generate a 3D parameter space \textbf{(c)} by convolving a sinusoid template \textbf{(b)} with the input transient sinogram. The peak of the parameter space represents the sinusoid parameters that best fit the input sinogram \textbf{(d)}.}
\label{fig:hough_transform}
\end{figure}

The key challenge is the presence of noise in a measured transient sinogram as shown in Figure~\ref{fig:hough_transform}(a). Similar to~\cite{sinusoid_hough}, we perform sinusoid fitting by using the Hough transform, which is commonly used as a robust parameter estimation approach. The Hough transform relies on a voting procedure to estimate the most likely parameters, which tends to be computationally expensive when there are many parameters to estimate. However, we can speed up the procedure in two ways. First, we can use a single sinusoid with a fixed temporal offset and phase shift as a template image for each amplitude (see Figure~\ref{fig:hough_transform}(b)).  Thus, we only need to prepare templates for the number of candidate amplitudes, which greatly reduces the computational cost. Second, we can also perform fast Hough transforms through a convolution, which can be efficiently computed in the Fourier domain. 
Here, we can define our sinusoidal 2D image template as
\begin{eqnarray}
    T_{\amplitude}(\theta, v) 
    = \left\{
    \begin{array}{rl}
    1 & \quad \text{if } v =  \amplitude \cos{(\theta)} + N/2\\
    0 & \quad \text{otherwise }
    \end{array} \right.
\end{eqnarray}
where $N$ represents temporal resolution of $T_{\amplitude}$. Convolving this 2D image template with the transient sinogram produces a slice of 2D parameter space image $A_{\amplitude}$:
\begin{eqnarray}
    A_{\amplitude} =  \mathit{\mathcal{F}^{-1}}( 
                        \mathit{\mathcal{F}}(T_{\amplitude}(\theta, v)) * 
                        \mathit{\mathcal{F}}(\tau(\theta, v))
                    )
\end{eqnarray}
where $\mathit{\mathcal{F}}, \mathit{\mathcal{F}^{-1}}$ are the Fourier and inverse Fourier transform operations.
Computing $A_{\amplitude}$ for each amplitude $\amplitude$ produces a three dimensional parameter space volume $A_{\{0:N/2\}}$. The element with the highest value indicates the sinusoidal parameters that best represents the sinusoid in the input transient.

\subsection{Object Localization with 3 Point Scanning}
\label{sec:3points}
\begin{figure}[h]
\begin{center}
\includegraphics[width=0.6\textwidth]{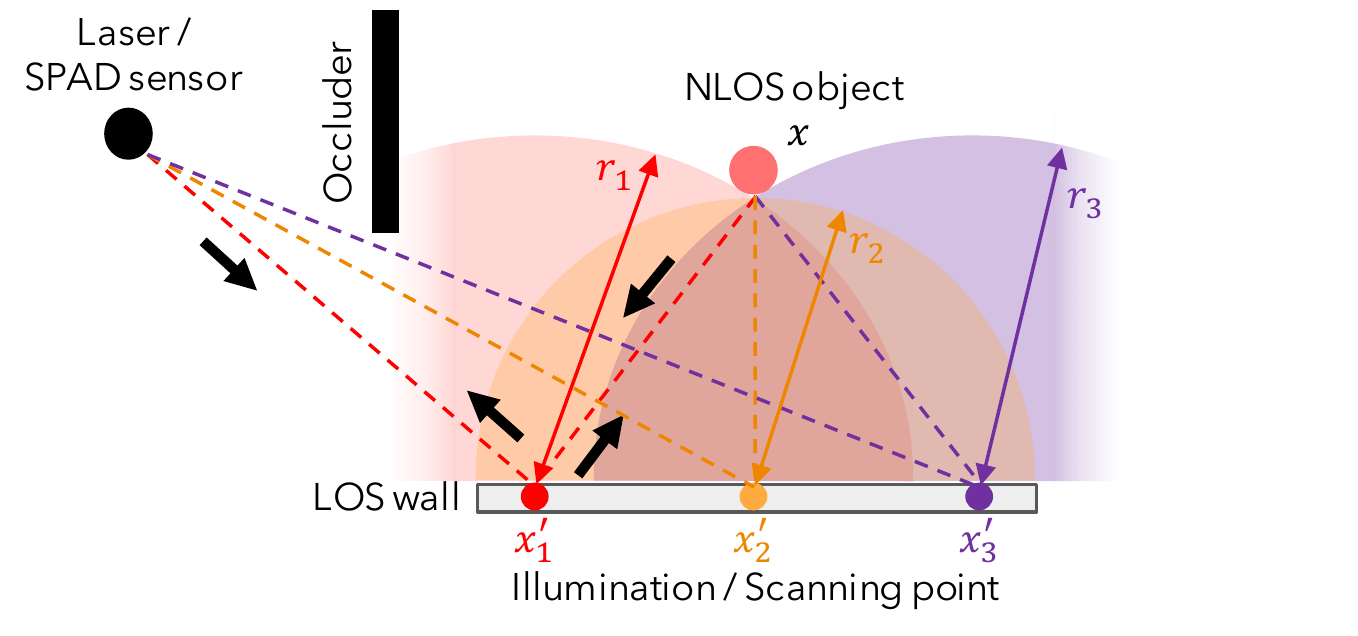}
\end{center}
\vspace{-10pt}
\caption{Object localization with just 3 scanning points~\cite{o2018confocal}}
\vspace{-2mm}
\label{fig:3points}
\end{figure}

\begin{figure}[htbp]
    \begin{subfigure}[]{0.32\linewidth}
		\centering
	    \includegraphics[width=1\linewidth]{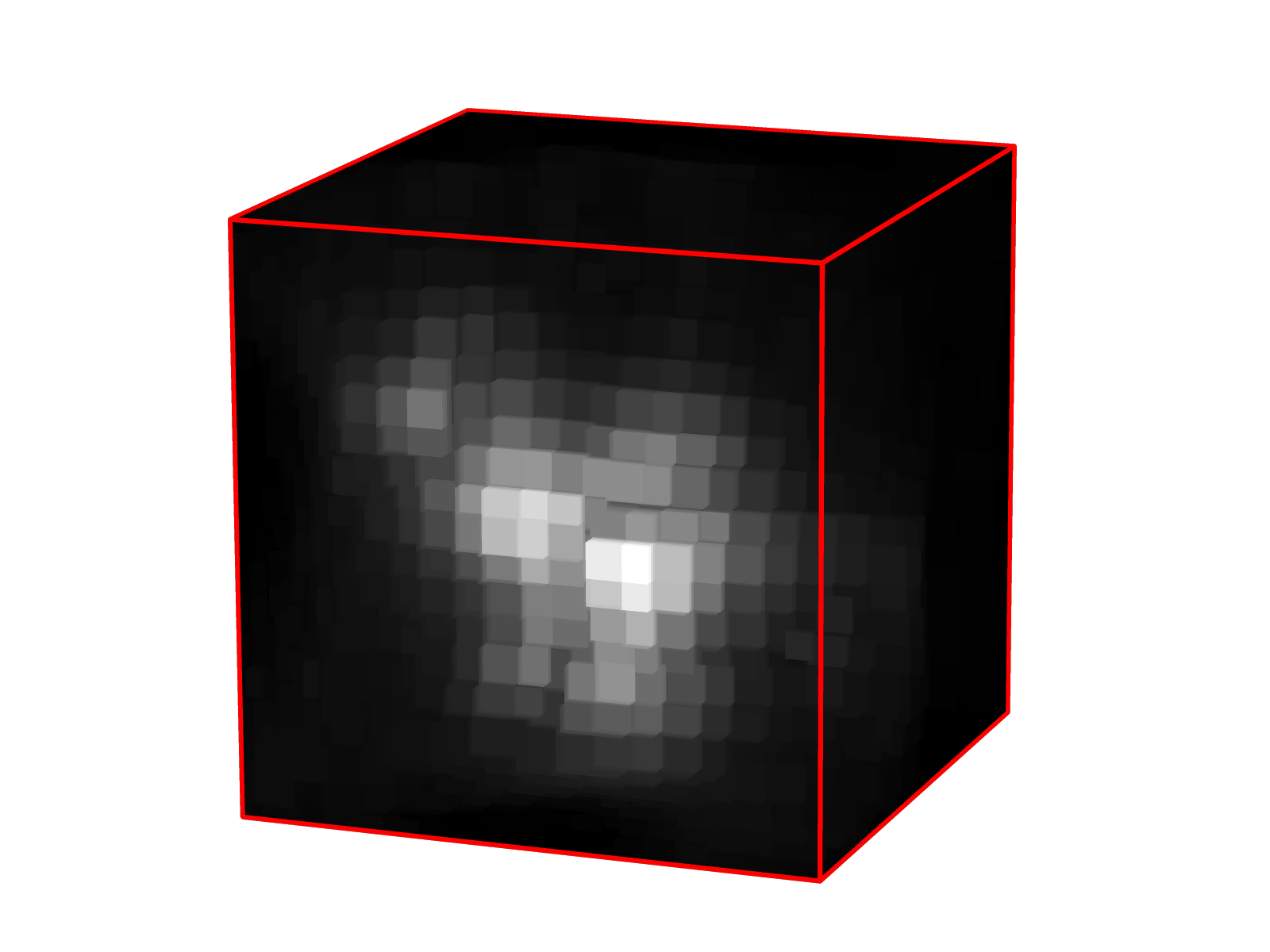}
	    \includegraphics[width=1\linewidth]{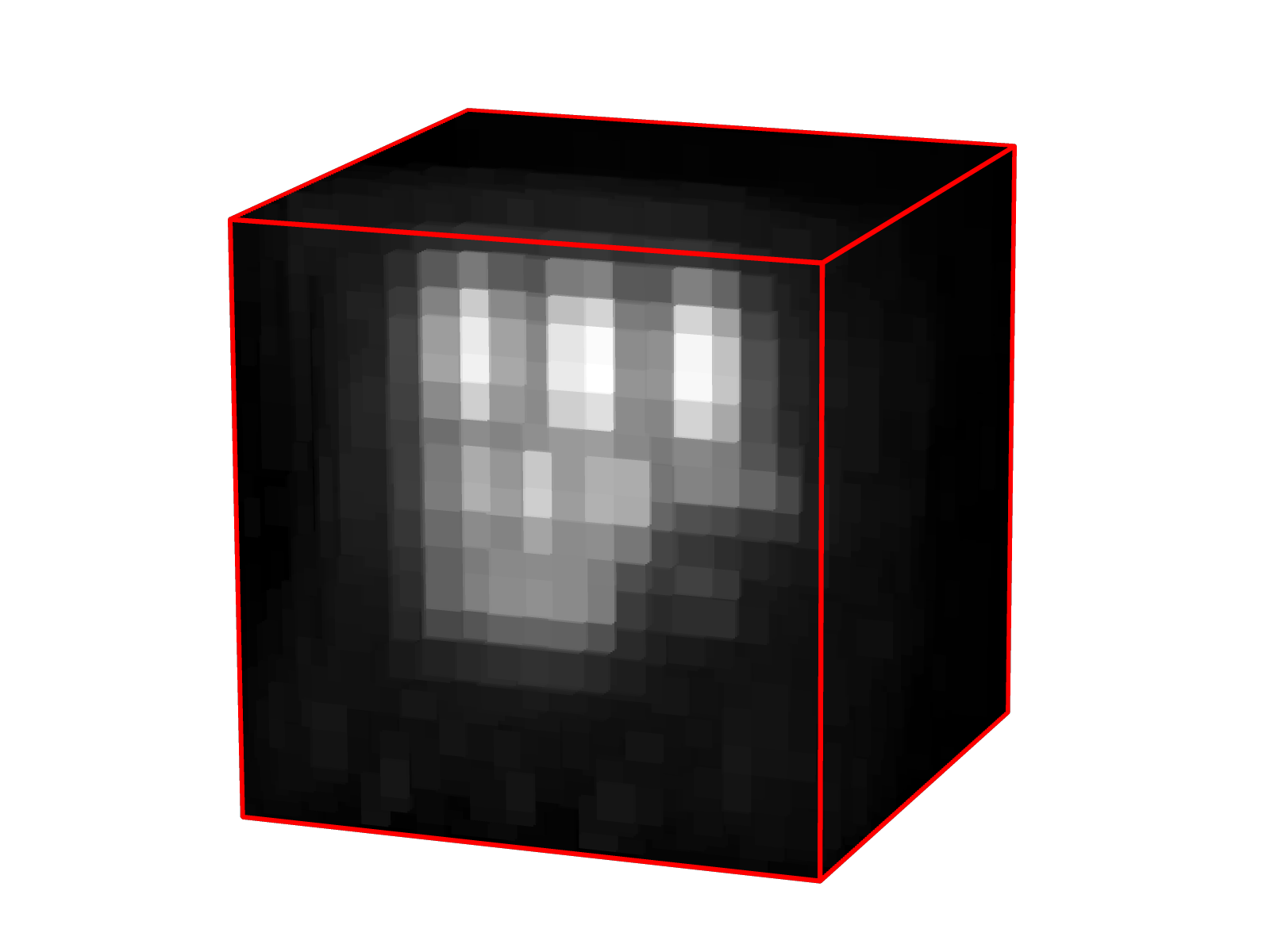}
	    \includegraphics[width=1\linewidth]{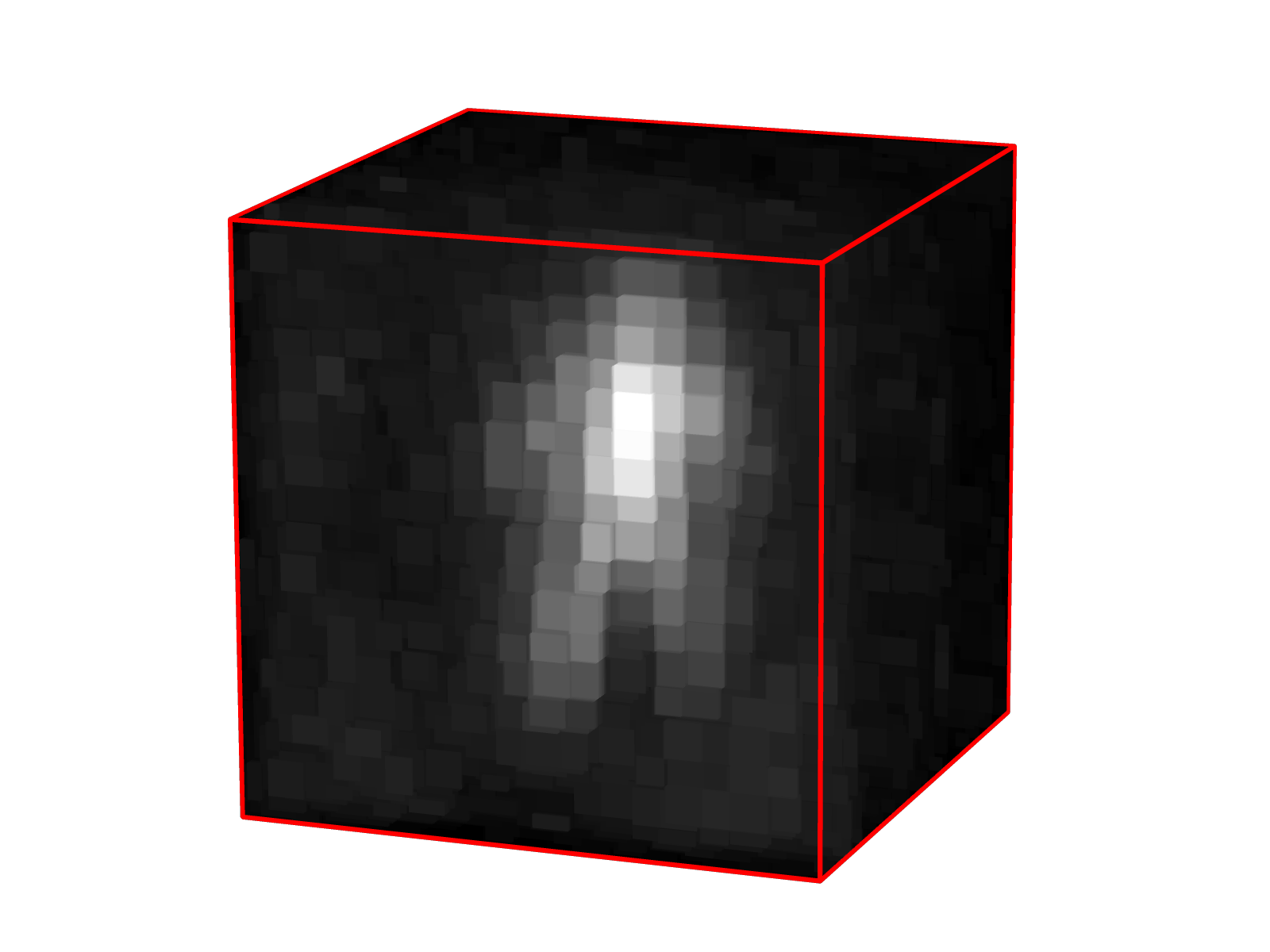}
	    \includegraphics[width=1\linewidth]{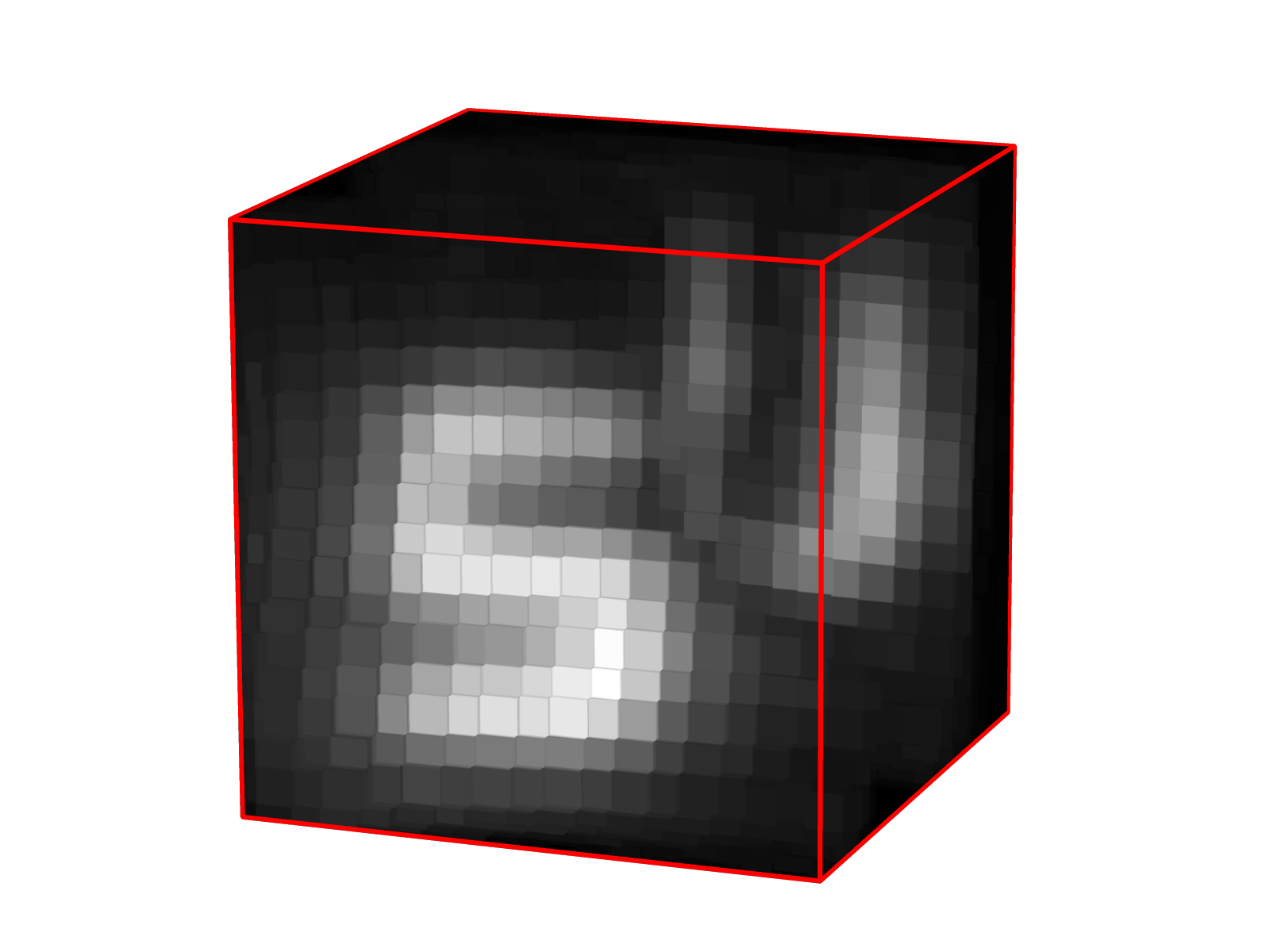}
		\caption{LCT}
	\end{subfigure}
	\begin{subfigure}[]{0.32\linewidth}
		\centering
	    \includegraphics[width=1\linewidth]{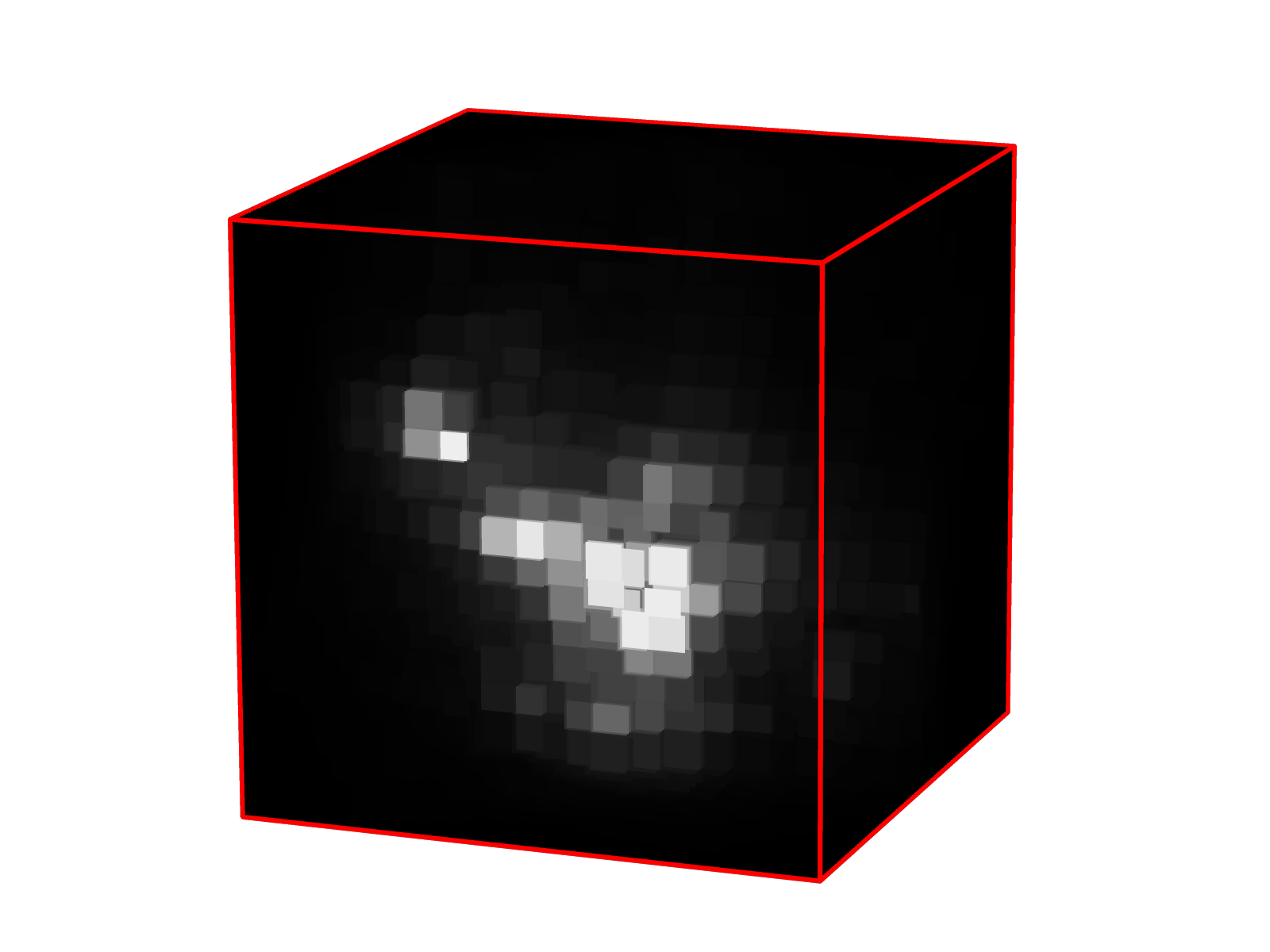}
	    \includegraphics[width=1\linewidth]{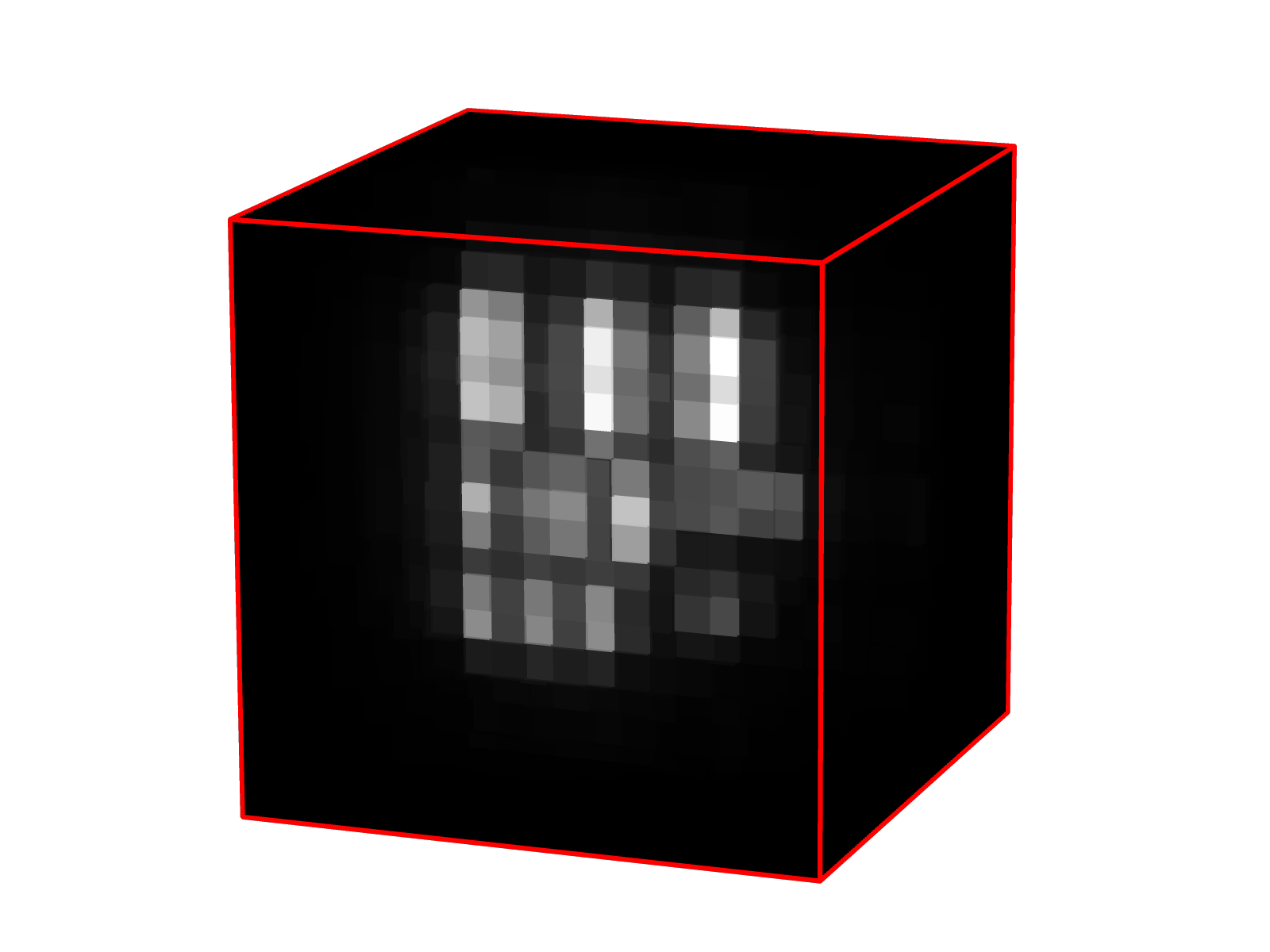}
	    \includegraphics[width=1\linewidth]{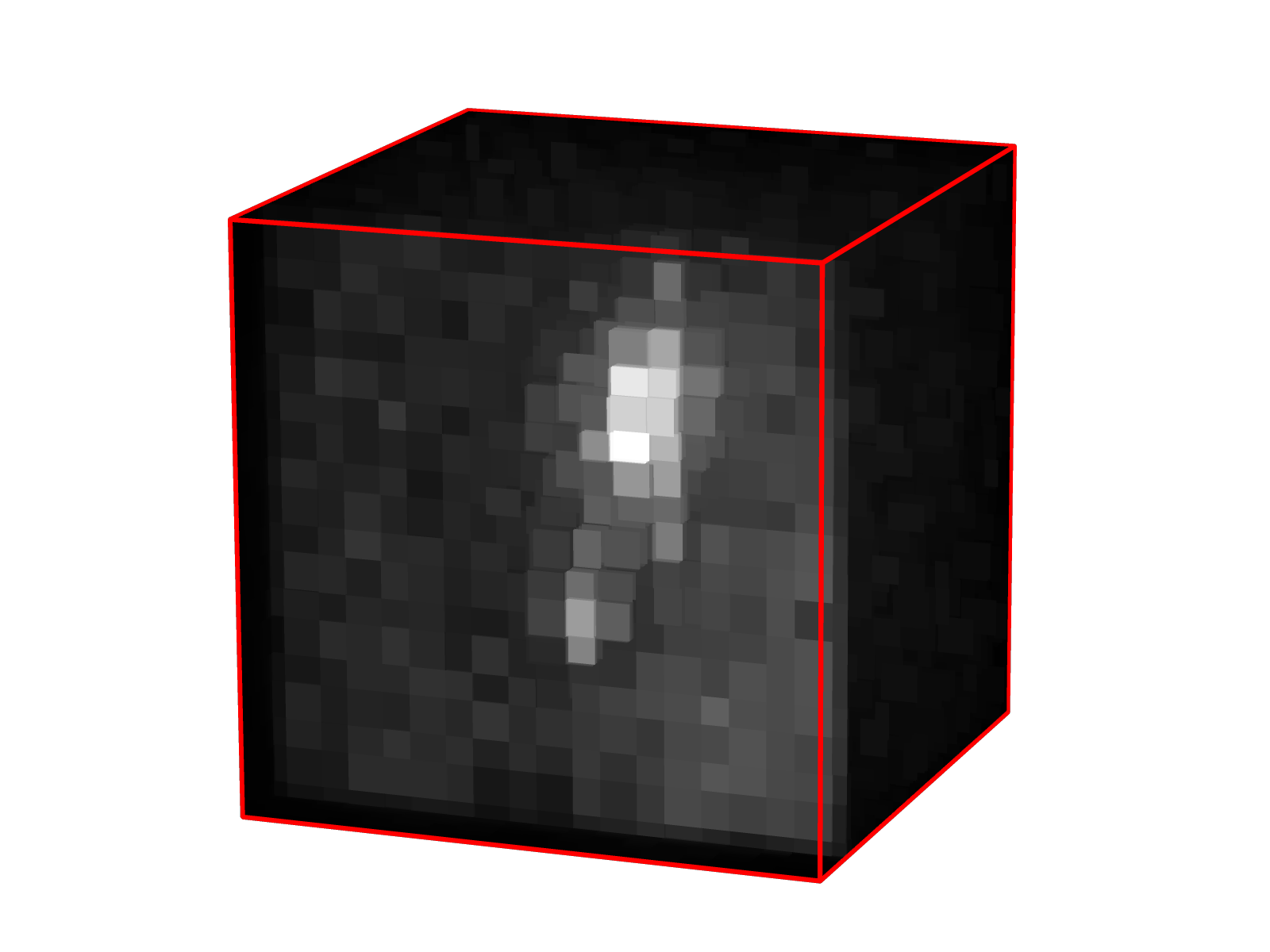}
	    \includegraphics[width=1\linewidth]{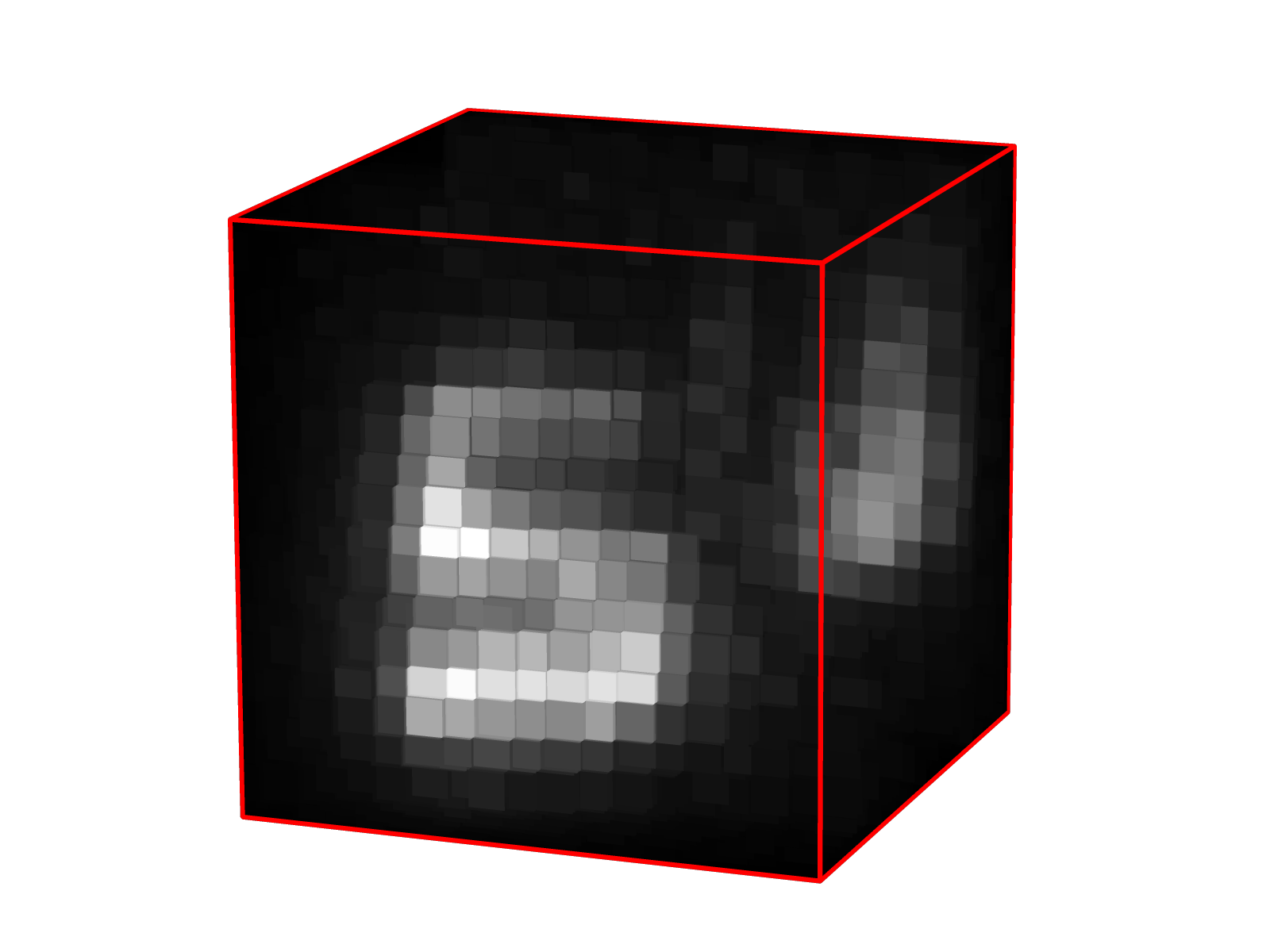}
		\caption{FK}
	\end{subfigure}
	\begin{subfigure}[]{0.32\linewidth}
		\centering
	    \includegraphics[width=1\linewidth]{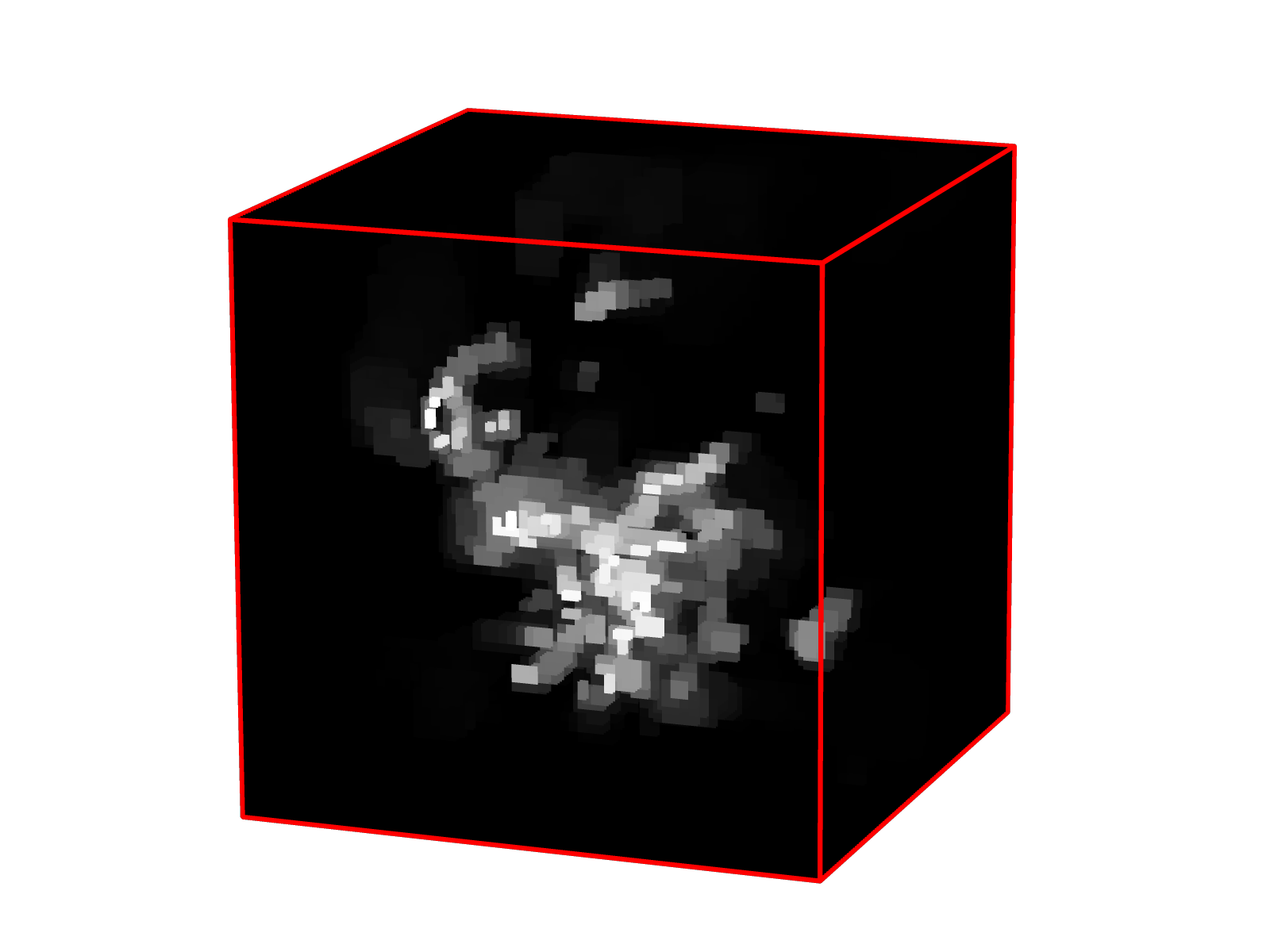}
	    \includegraphics[width=1\linewidth]{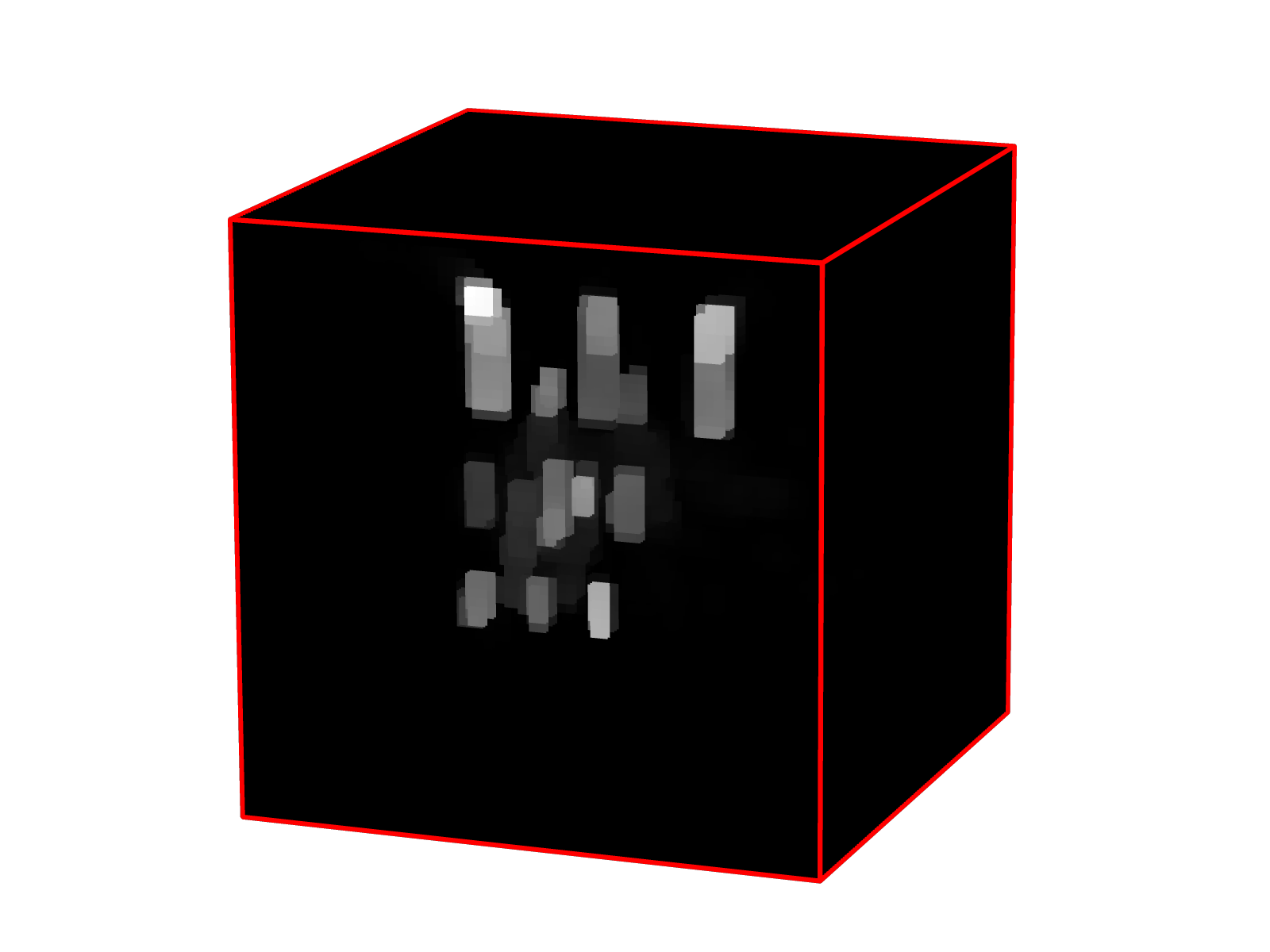}
	    \includegraphics[width=1\linewidth]{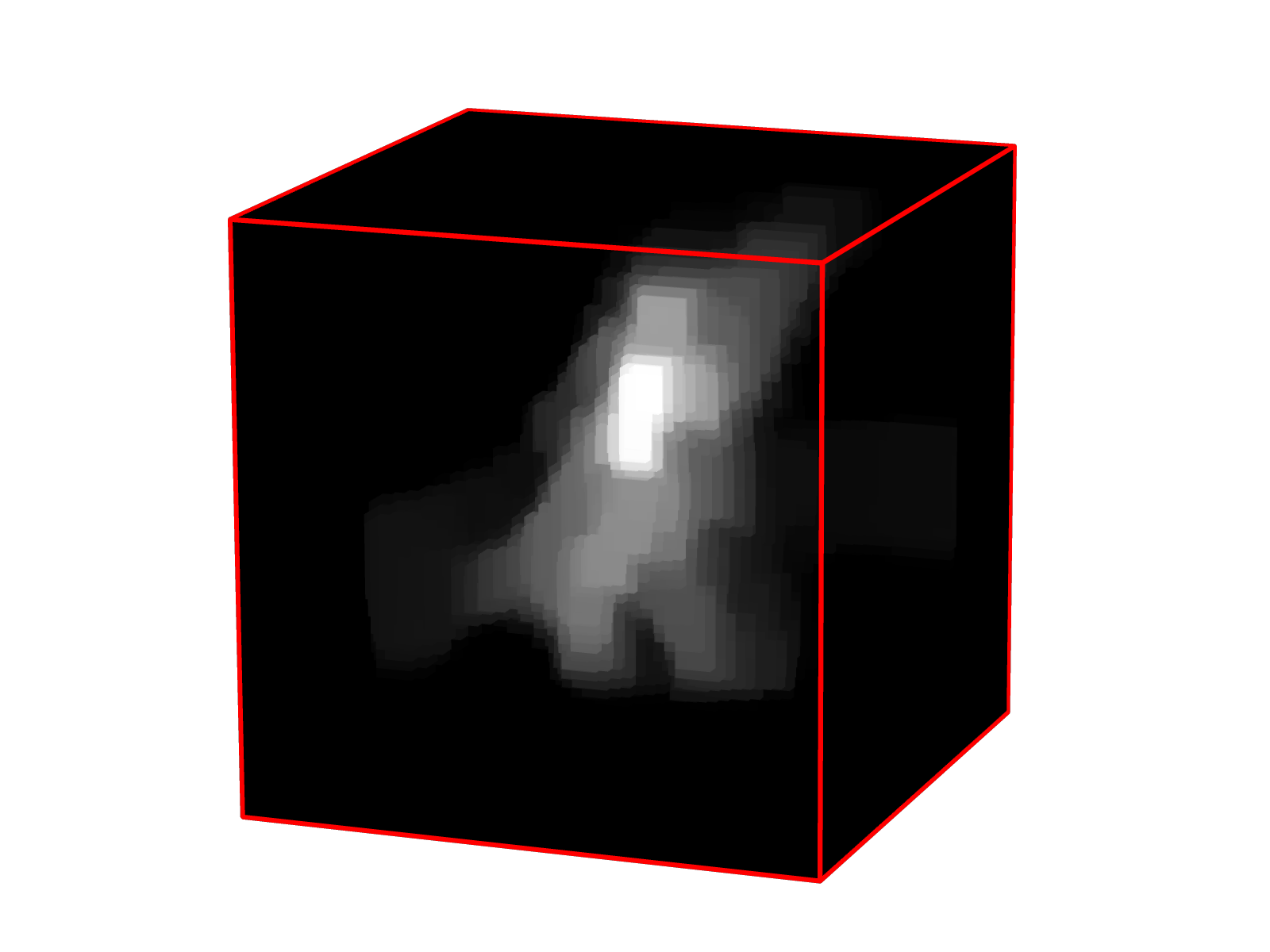}
	    \includegraphics[width=1\linewidth]{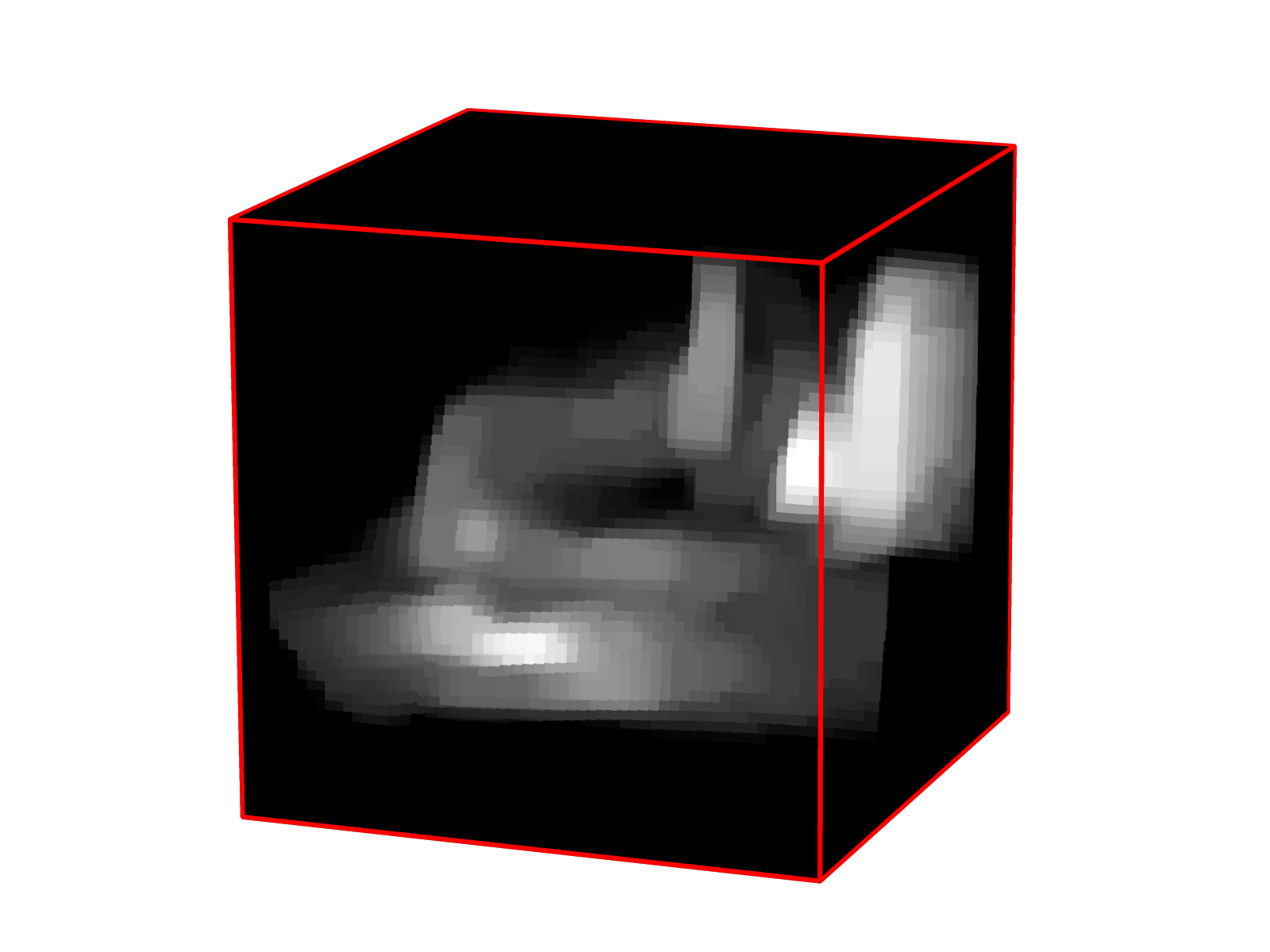}
		\caption{\CCNLOS}
	\end{subfigure}
	\caption{Qualitative comparison of 3D reconstruction quality for a fixed budget of 360 samples. For FK/LCT, we used a $19 \times 19$ grid. In this scenario, while both scanning patterns use the same number of samples, a \CCNLOS pattern can be acquired 19 times faster.}
	\label{fig:3d_19}
\end{figure}

We used 3 scanning point trilateration~\cite{o2018confocal} as a baseline in Section~5.1 for single object localization. We describe this procedure in further detail in this section. 
As illustrated in Figure~\ref{fig:3points}, let $x$ denote the position of the NLOS object, and $x'_{1}$, $x'_{2}$, and $x'_{3}$ the positions of each scanning point. Assuming that there is only a single object in the hidden scene, the temporal peaks of the transient measurements for each scanning point $t_{1}, t_{2}, t_{3}$ directly give the distance between the scanning points and the hidden object. These distances can be calculated from the temporal peaks as $r_{1}=t_{1}c/2$, $r_{2}=t_{2}c/2$, $r_{3}=t_{3}c/2$, where $c$ denotes the speed of light. 
Since the object exists at the intersection of the 3 spheres centered at $x'_{1}$, $x'_{2}$, and $x'_{3}$ with radius $r_{1}, r_{2}$, and $r_{3}$ respectively, the object position $x$ is obtained by solving the three following simultaneous equations:

\begin{equation}
(x'_{1} - x)^2 = r_{1}^2
\end{equation}
\begin{equation}
(x'_{2} - x)^2 = r_{2}^2
\end{equation}
\begin{equation}
(x'_{3} - x)^2 = r_{3}^2
\end{equation}

\subsection{3D Volume Imaging via ADMM}
\label{sec:admm}

As mentioned in the main paper, we recover a full 3D volume reconstruction $\bm{\rho}$ of the hidden scene from \CCNLOS measurements $\bm{\tau}_{\text{circ}}$, by using a modified version of the iterative light cone transform (LCT) procedure used in confocal NLOS imaging~\cite{o2018confocal}. We decide to use non-negativity, sparsity, and total variation priors to compensate for the very sparsely sampled input data. With these priors, the optimization problem can be written as:

\begin{equation}
\scriptstyle
    \min_{\bm{\rho}} \quad \frac{1}{2} \left\Vert \bm{\tau}_{\text{circ}} - \mathbf{M}\mathbf{A}\bm{\rho} \right\Vert ^2_2 + \mathcal{I}_{\mathbb{R}_+}(\bm{\rho}) + \lambda_s \left\Vert \bm{\rho} \right \Vert_1 + \lambda_{TV} \left( \left\Vert \mathbf{D}_x \bm{\rho} \right \Vert_1 + \left\Vert \mathbf{D}_y \bm{\rho} \right \Vert_1 + \left\Vert \mathbf{D}_z \bm{\rho} \right \Vert_1\right)
\end{equation}
For simplicity, we modify the above optimization to operate in the light cone domain~\cite{o2018confocal}:

\begin{equation}
\scriptstyle
    \min_{\bm{\rho}_{u}} \quad \frac{1}{2} \left\Vert \bm{\tau}_{\text{circ}} - \mathbf{M}\mathbf{H}\bm{\rho}_{u} \right\Vert ^2_2 + \mathcal{I}_{\mathbb{R}_+}(\bm{\rho}_{u}) + \lambda_s \left\Vert \bm{\rho}_u \right \Vert_1 + \lambda_{TV} \left( \left\Vert \mathbf{D}_x \bm{\rho}_u \right \Vert_1 + \left\Vert \mathbf{D}_y \bm{\rho}_u \right \Vert_1 + \left\Vert \mathbf{D}_z \bm{\rho}_u \right \Vert_1\right)
\end{equation}

where $\textbf{H}$ is a convolutional matrix, $\bm{\rho}_u$ is a resampled version of $\bm{\rho}$ following the LCT procedure, and $\mathbf{D}_x$, $\mathbf{D}_y$, and $\mathbf{D}_z$ implement finite difference operators along the $x$, $y$, and $z$ directions respectively. For notational simplicity, let $\bm{\rho} = \bm{\rho}_u$ and $\bm{\tau} = \bm{\tau}_{\text{circ}}$. In order to apply ADMM, we can rewrite the above equation:

\small
\begin{align}
    \min_{\bm{\rho}} \quad & \underbrace{\frac{1}{2} \left\Vert \bm{\tau} - \mathbf{z}_1 \right\Vert ^2_2}_{g_1(\mathbf{z}_1)} + \underbrace{\mathcal{I}_{\mathbb{R}_+}(\mathbf{z}_2)}_{g_2(\mathbf{z}_2)} + \underbrace{\lambda_s \left\Vert \mathbf{z}_3 \right \Vert_1}_{g_3(\mathbf{z}_3)} + \underbrace{\lambda_{TV} \left\Vert \mathbf{z}_4 \right \Vert_1}_{g_4(\mathbf{z}_4)} + \underbrace{\lambda_{TV} \left\Vert \mathbf{z}_5 \right \Vert_1}_{g_5(\mathbf{z}_5)} + \underbrace{\lambda_{TV} \left\Vert \mathbf{z}_6 \right \Vert_1}_{g_6(\mathbf{z}_6)}\\
    \textrm{s.t.} \quad & \underbrace{\begin{bmatrix}\mathbf{M}\mathbf{H}\\\mathbf{I}\\\mathbf{I}\\\mathbf{D}_x\\\mathbf{D}_y\\\mathbf{D}_z\end{bmatrix}}_\mathbf{C} \bm{\rho} - \underbrace{\begin{bmatrix}\mathbf{z}_1\\\mathbf{z}_2\\\mathbf{z}_3\\\mathbf{z}_4\\\mathbf{z}_5\\\mathbf{z}_6\end{bmatrix}}_\mathbf{z}=\mathbf{0}
\end{align}
\normalsize

The Augmented Lagrangian for this objective function can be written as:

\begin{equation}
    L_\mu(\bm{\rho}, \mathbf{z}, \mathbf{y}) = \sum_{i=1}^{6}g_i(\mathbf{z}_i) + \mathbf{y}^T(\mathbf{C}\bm{\rho} - \mathbf{z}) + \frac{\mu}{2} \left\Vert \mathbf{C}\bm{\rho} - \mathbf{z} \right\Vert^2_2
\end{equation}

We operate on the scaled form, with $\mathbf{u} = \mathbf{y} / \mu$:

\begin{equation}
    L_\mu(\bm{\rho}, \mathbf{z}, \mathbf{u}) = \sum_{i=1}^{6}g_i(\mathbf{z}_i) + \frac{\mu}{2} \left\Vert \mathbf{C}\bm{\rho} - \mathbf{z} + \mathbf{u} \right\Vert^2_2 - \frac{\mu}{2} \left\Vert \mathbf{u} \right\Vert^2_2
\end{equation}
We are now ready to write out the proximal operator update rules:
\begin{align*}
    \mathbf{z}_1 &\leftarrow \argmin_{\mathbf{z}_1} g_1(\mathbf{z}_1) 
    + \frac{\mu}{2} \left\Vert \mathbf{v} - \mathbf{z}_1 \right\Vert^2_2, \quad \mathbf{v} = \mathbf{M} \mathbf{H}\bm{\rho}_u + \mathbf{u}_1\\
    &= \argmin_{\mathbf{z}_1} \frac{1}{2}\left\Vert \bm{\tau} - \mathbf{z}_1\right\Vert^2_2
    + \frac{\mu}{2} \left\Vert \mathbf{v} - \mathbf{z}_1 \right\Vert^2_2\\
    &= \frac{\bm{\tau} + \mu \mathbf{v}}{1 + \mu}\\
    \mathbf{z}_2 &\leftarrow \argmin_{\mathbf{z}_2} g_2(\mathbf{z}_2) 
    + \frac{\mu}{2} \left\Vert \mathbf{v} - \mathbf{z}_2 \right\Vert^2_2, \quad \mathbf{v} = \bm{\rho} + \mathbf{u}_2\\
    &= \argmin_{\mathbf{z}_2} \mathcal{I}_{\mathbb{R}_+}(\bm{z}_2) + \frac{\mu}{2} \left\Vert \mathbf{v} - \mathbf{z}_2 \right\Vert^2_2\\
    &= \text{max}(0, \mathbf{v})\\
    \mathbf{z}_3 &\leftarrow \argmin_{\mathbf{z}_3} g_3(\mathbf{z}_3) 
    + \frac{\mu}{2} \left\Vert \mathbf{v} - \mathbf{z}_3 \right\Vert^2_2, \quad \mathbf{v} = \bm{\rho} + \mathbf{u}_3\\
    &= \argmin_{\mathbf{z}_3} \lambda_s \left\Vert \bm{z}_3 \right \Vert_1  + \frac{\mu}{2} \left\Vert \mathbf{v} - \mathbf{z}_3 \right\Vert^2_2\\
    &= S_{\lambda_s/\mu} (\mathbf{v})\\
    \mathbf{z}_4 &\leftarrow \argmin_{\mathbf{z}_4} g_4(\mathbf{z}_4) 
    + \frac{\mu}{2} \left\Vert \mathbf{v} - \mathbf{z}_4 \right\Vert^2_2, \quad \mathbf{v} = \mathbf{D}_x \bm{\rho} + \mathbf{u}_4\\
    &= \argmin_{\mathbf{z}_4} \lambda_{TV} \left\Vert \bm{z}_4 \right \Vert_1 + \frac{\mu}{2} \left\Vert \mathbf{v} - \mathbf{z}_4 \right\Vert^2_2\\
    &= S_{\lambda_{TV}/\mu} (\mathbf{v})\\
    \mathbf{z}_5 &\leftarrow \argmin_{\mathbf{z}_5} g_5(\mathbf{z}_5) 
    + \frac{\mu}{2} \left\Vert \mathbf{v} - \mathbf{z}_5 \right\Vert^2_2, \quad \mathbf{v} = \mathbf{D}_y \bm{\rho} + \mathbf{u}_5\\
    &= \argmin_{\mathbf{z}_5} \lambda_{TV} \left\Vert \bm{z}_5 \right \Vert_1  + \frac{\mu}{2} \left\Vert \mathbf{v} - \mathbf{z}_5 \right\Vert^2_2\\
    &= S_{\lambda_{TV}/\mu} (\mathbf{v})\\
    \mathbf{z}_6 &\leftarrow \argmin_{\mathbf{z}_6} g_6(\mathbf{z}_6) 
    + \frac{\mu}{2} \left\Vert \mathbf{v} - \mathbf{z}_6 \right\Vert^2_2, \quad \mathbf{v} = \mathbf{D}_z \bm{\rho} + \mathbf{u}_6\\
    &= \argmin_{\mathbf{z}_6} \lambda_{TV} \left\Vert \bm{z}_6 \right \Vert_1  + \frac{\mu}{2} \left\Vert \mathbf{v} - \mathbf{z}_6 \right\Vert^2_2\\
    &= S_{\lambda_{TV}/\mu} (\mathbf{v})\\
    \mathbf{u} &\leftarrow \mathbf{u} + \mathbf{C}\bm{\rho} - \mathbf{z}\\
    \bm{\rho} &\leftarrow \argmin_{\bm{\rho}} \frac{1}{2} \left\Vert \mathbf{C} \bm{\rho} - \mathbf{v}\right\Vert ^2_2, \quad \mathbf{v} = \mathbf{z} - \mathbf{u} \\
\end{align*}
It is difficult to solve the proximal operator for $\bm{\rho}$ in closed form with the lossy mapping term $\mathbf{M}$. Instead, we opt for a linearized ADMM approach. Our update rule for $\bm{\rho}$ now looks like the following:
\begin{equation}
    \bm{\rho} \leftarrow \bm{\rho} - \frac{\mu}{\nu}\mathbf{C}^*(\mathbf{C}\bm{\rho} - \mathbf{v})
\end{equation}
where $\nu$ controls the learning rate. Under this formulation, all of the above proximal operators can be efficiently solved, because $\mathbf{H}$, $\mathbf{D}_x$, $\mathbf{D}_y$, and $\mathbf{D}_z$ can all be expressed as elementwise multiplications in the Fourier domain.

\begin{figure}[tb]
    \begin{subfigure}[]{0.32\linewidth}
		\centering
	    \includegraphics[width=1\linewidth]{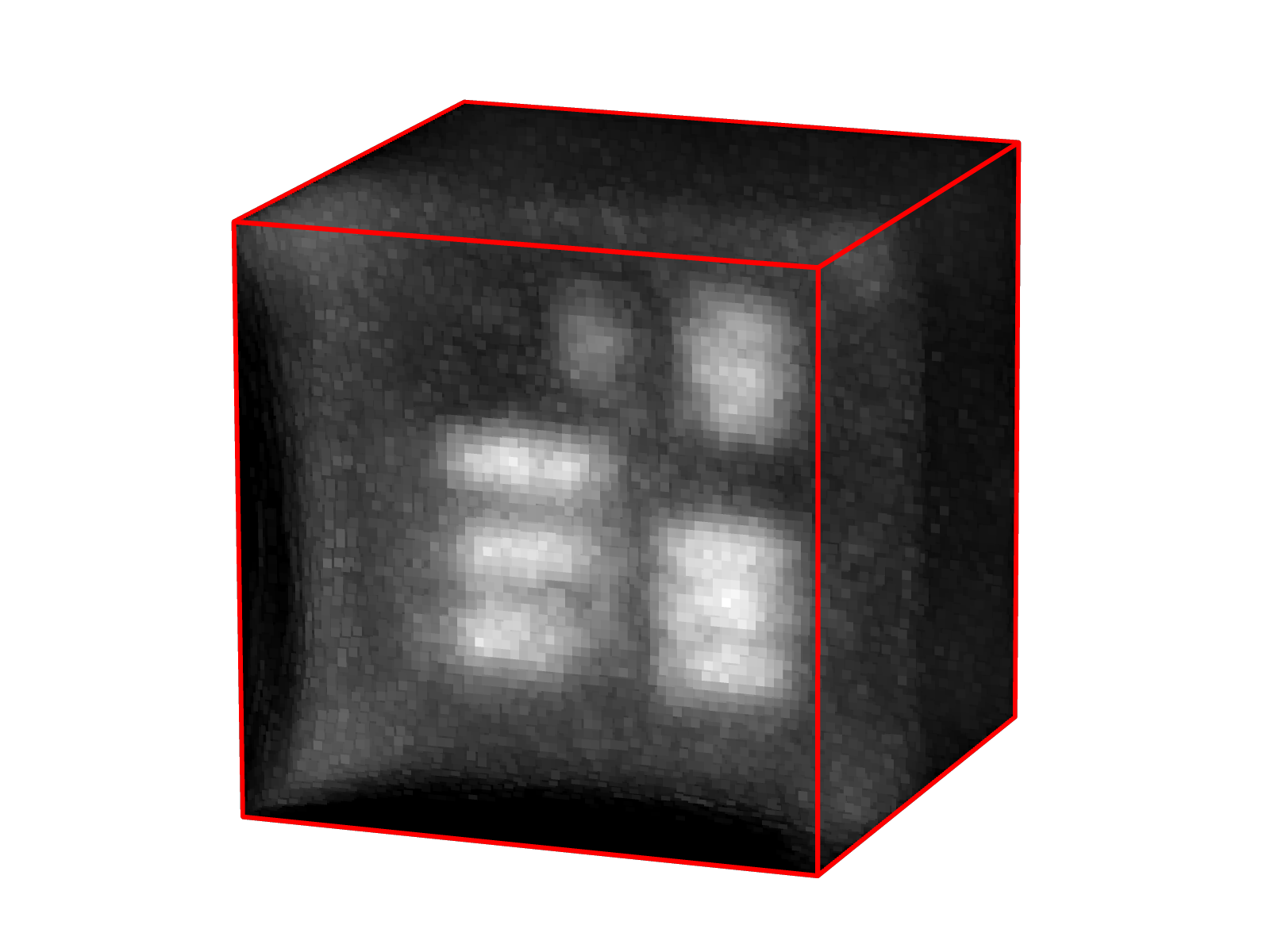}
	    \includegraphics[width=1\linewidth]{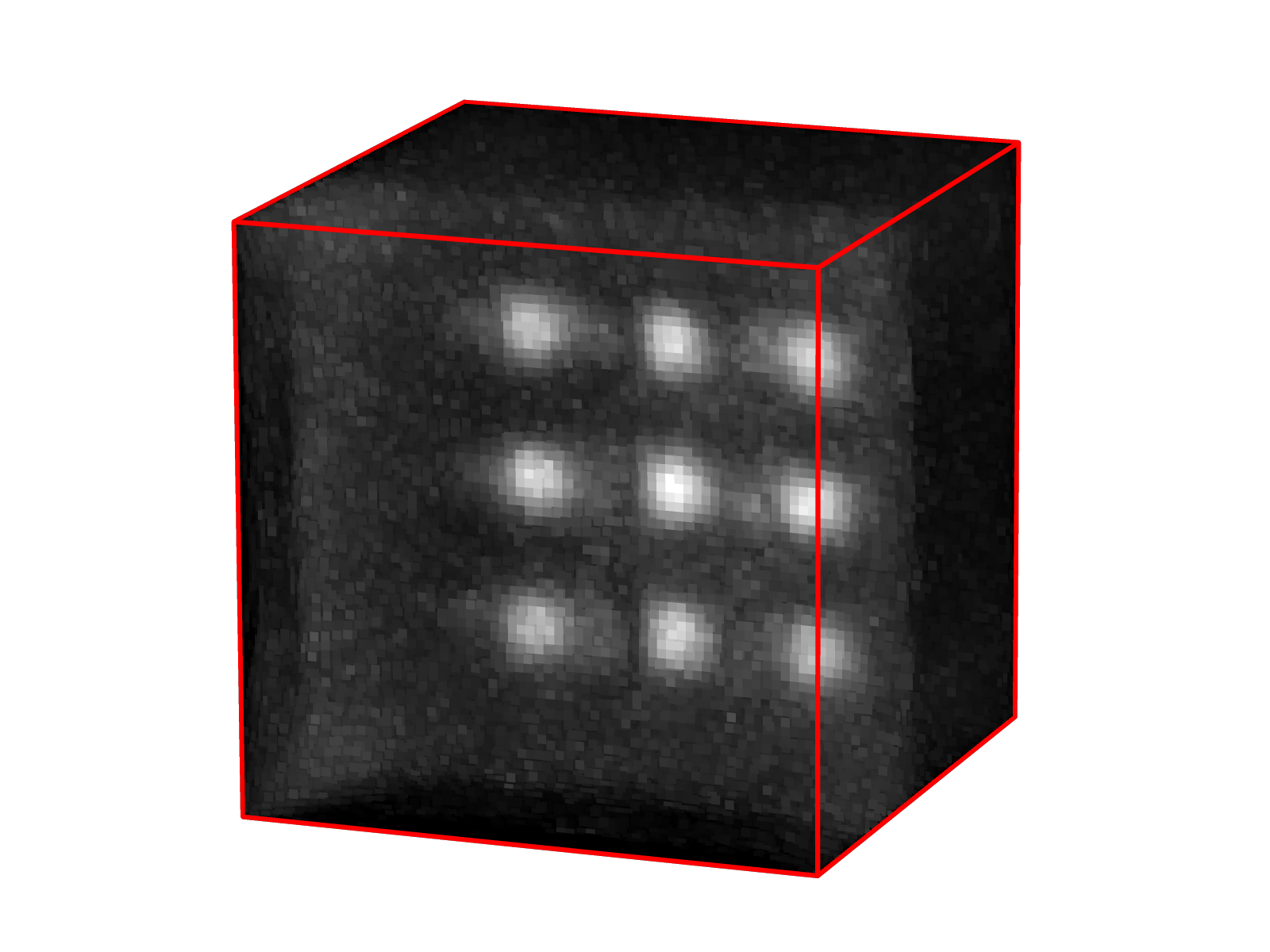}
		\caption{LCT}
	\end{subfigure}
	\begin{subfigure}[]{0.32\linewidth}
		\centering
	    \includegraphics[width=1\linewidth]{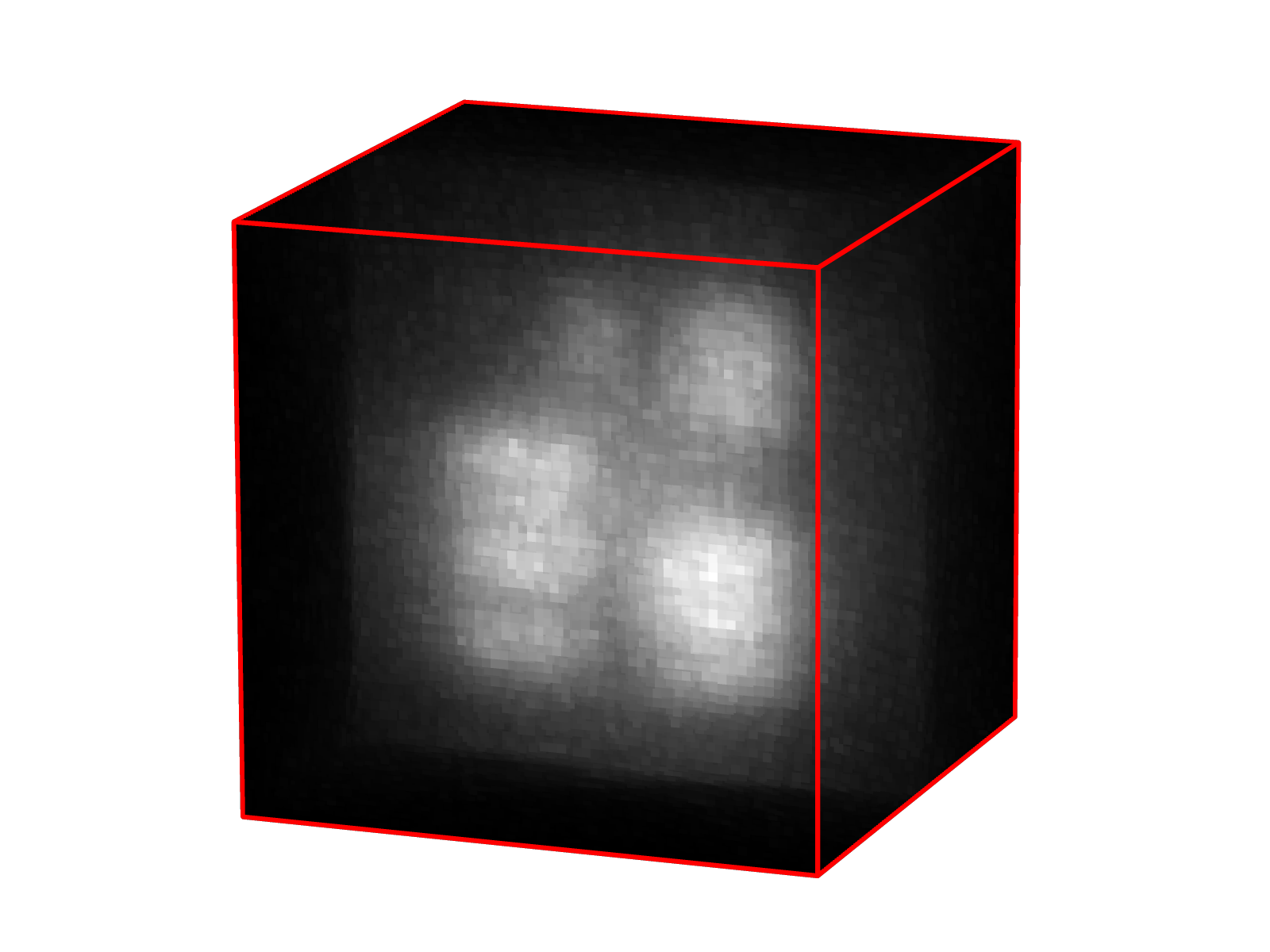}
	    \includegraphics[width=1\linewidth]{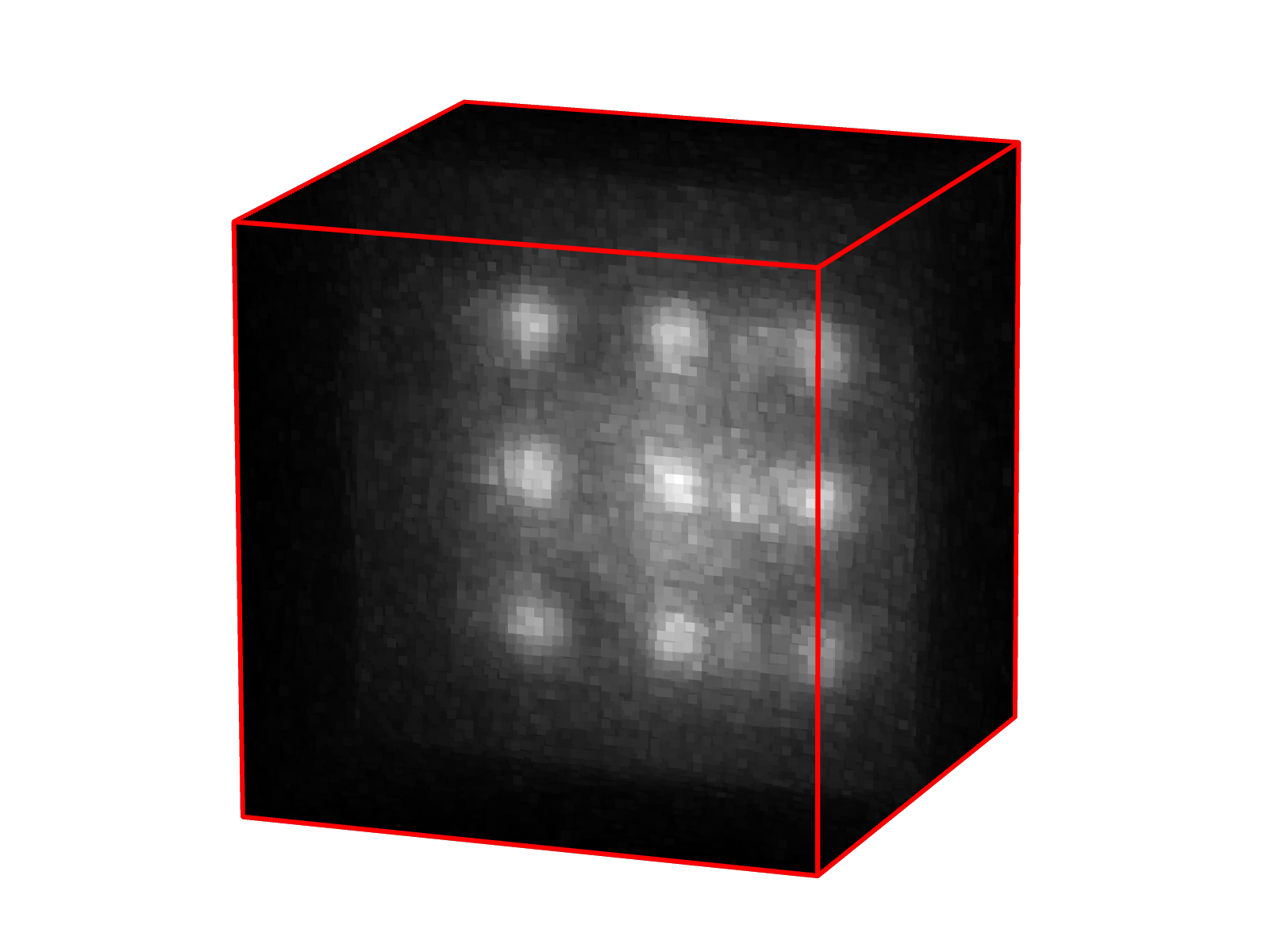}
		\caption{FK}
	\end{subfigure}
	\begin{subfigure}[]{0.32\linewidth}
		\centering
	    \includegraphics[width=1\linewidth]{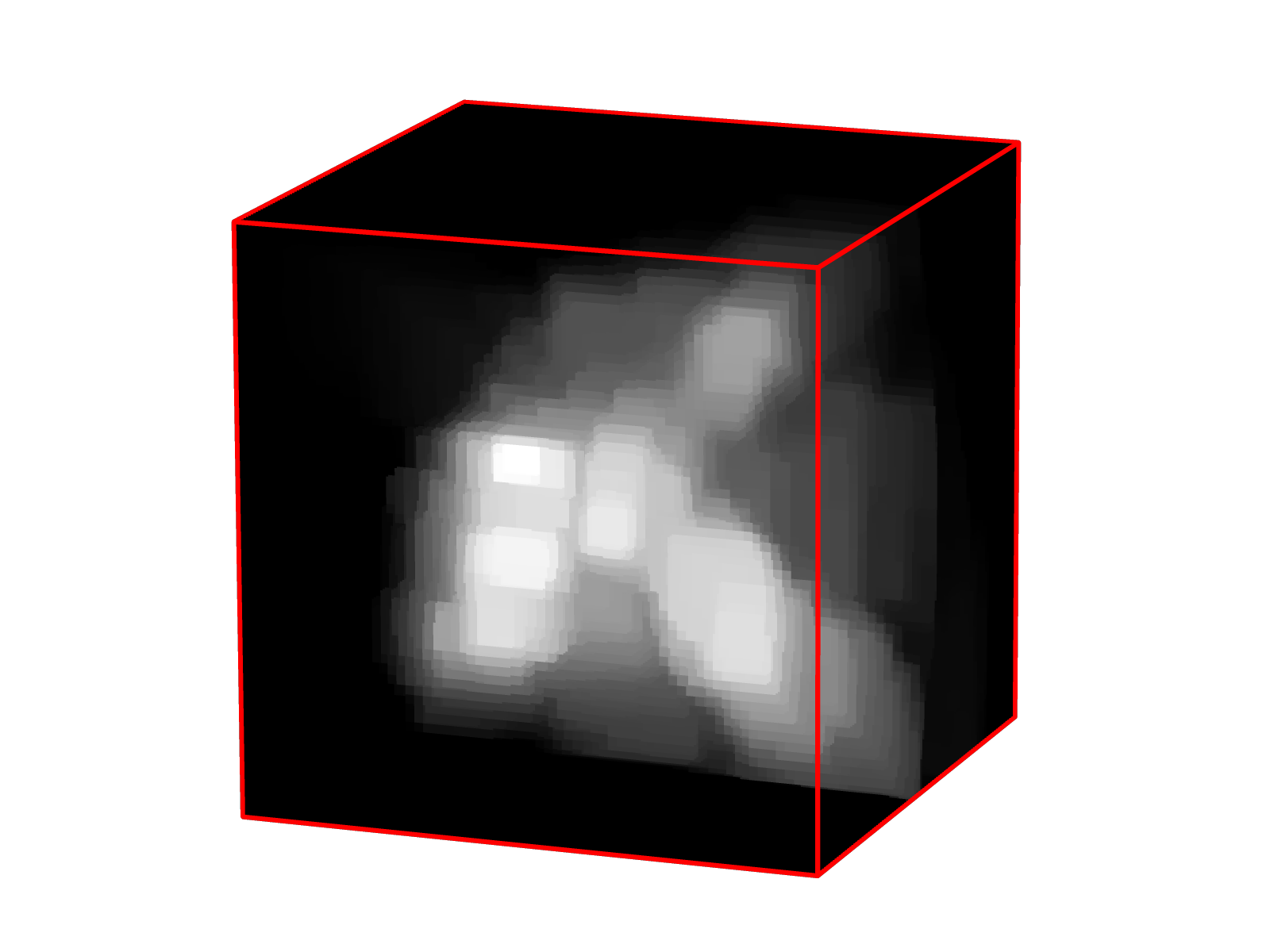}
	    \includegraphics[width=1\linewidth]{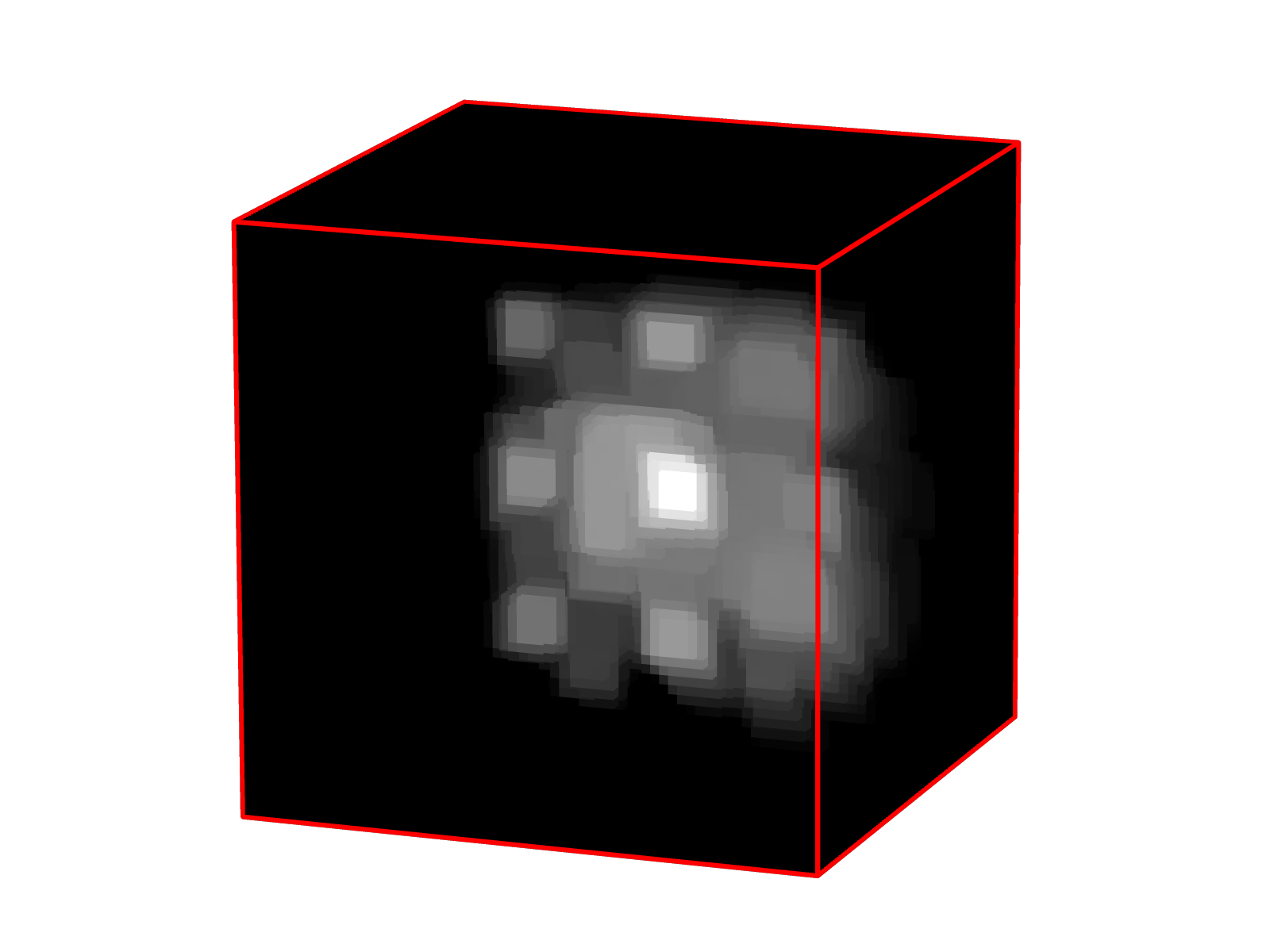}
		\caption{\CCNLOS}
	\end{subfigure}
	\caption{3D reconstruction examples on real data from O'Toole \etal~\cite{o2018confocal}. Our circular scans contain significant information about the hidden scene, capturing the important features of the hidden scene. }
	\label{fig:3d_real}
\end{figure}

\begin{figure}[h!]
    \begin{subfigure}[]{0.32\linewidth}
		\centering
	    \includegraphics[width=1\linewidth]{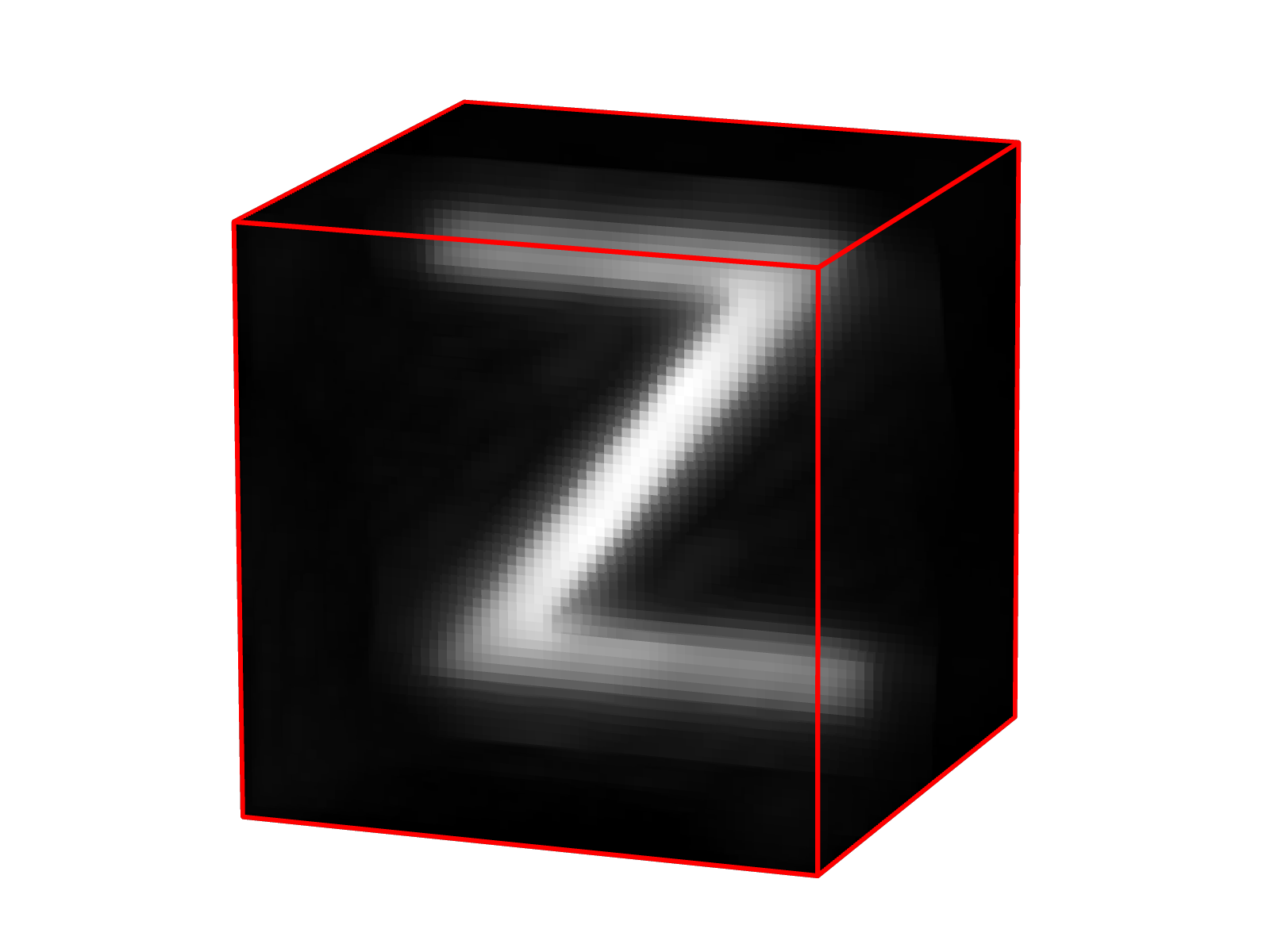}
	    \includegraphics[width=1\linewidth]{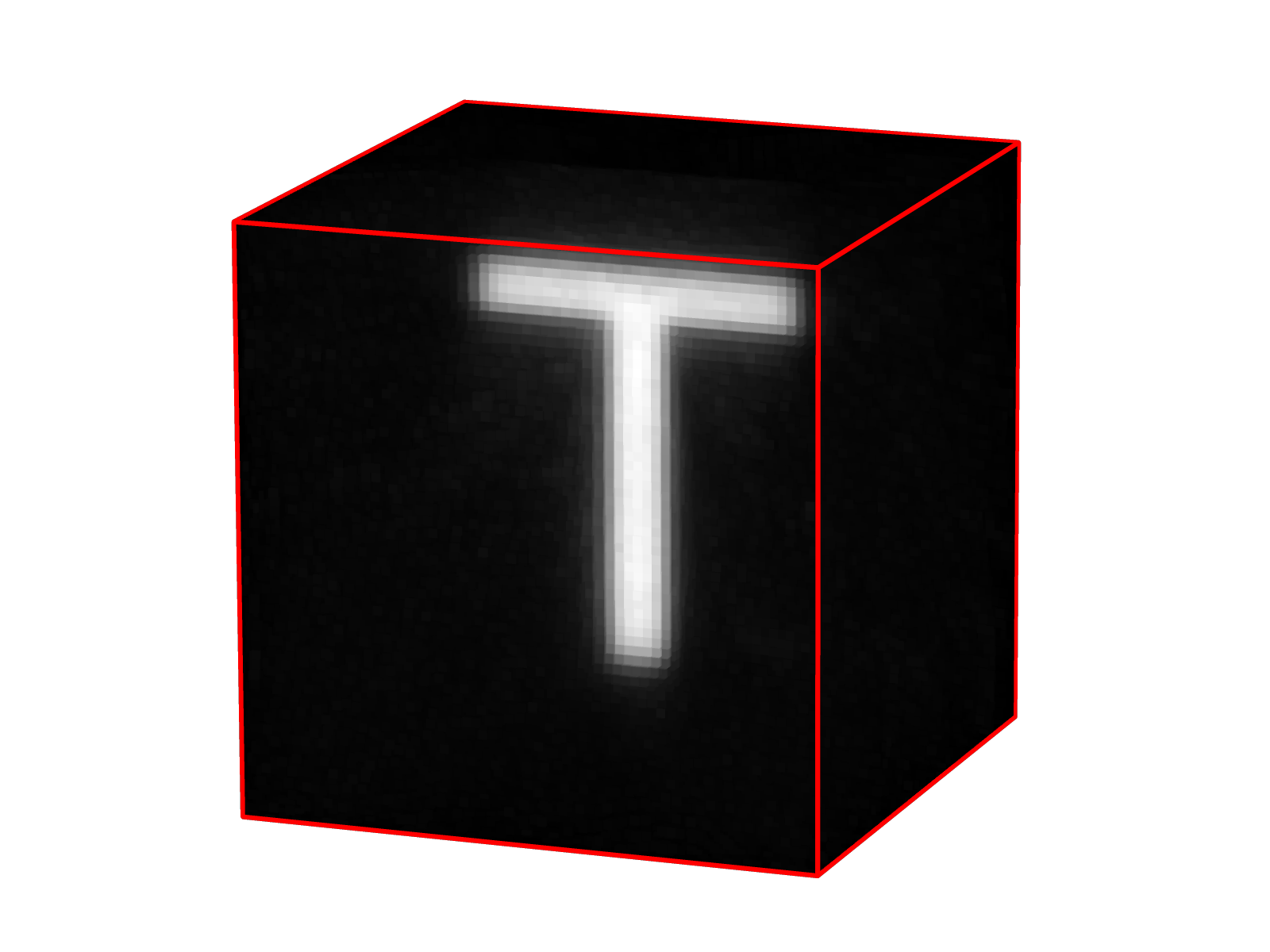}
		\caption{LCT}
	\end{subfigure}
	\begin{subfigure}[]{0.32\linewidth}
		\centering
	    \includegraphics[width=1\linewidth]{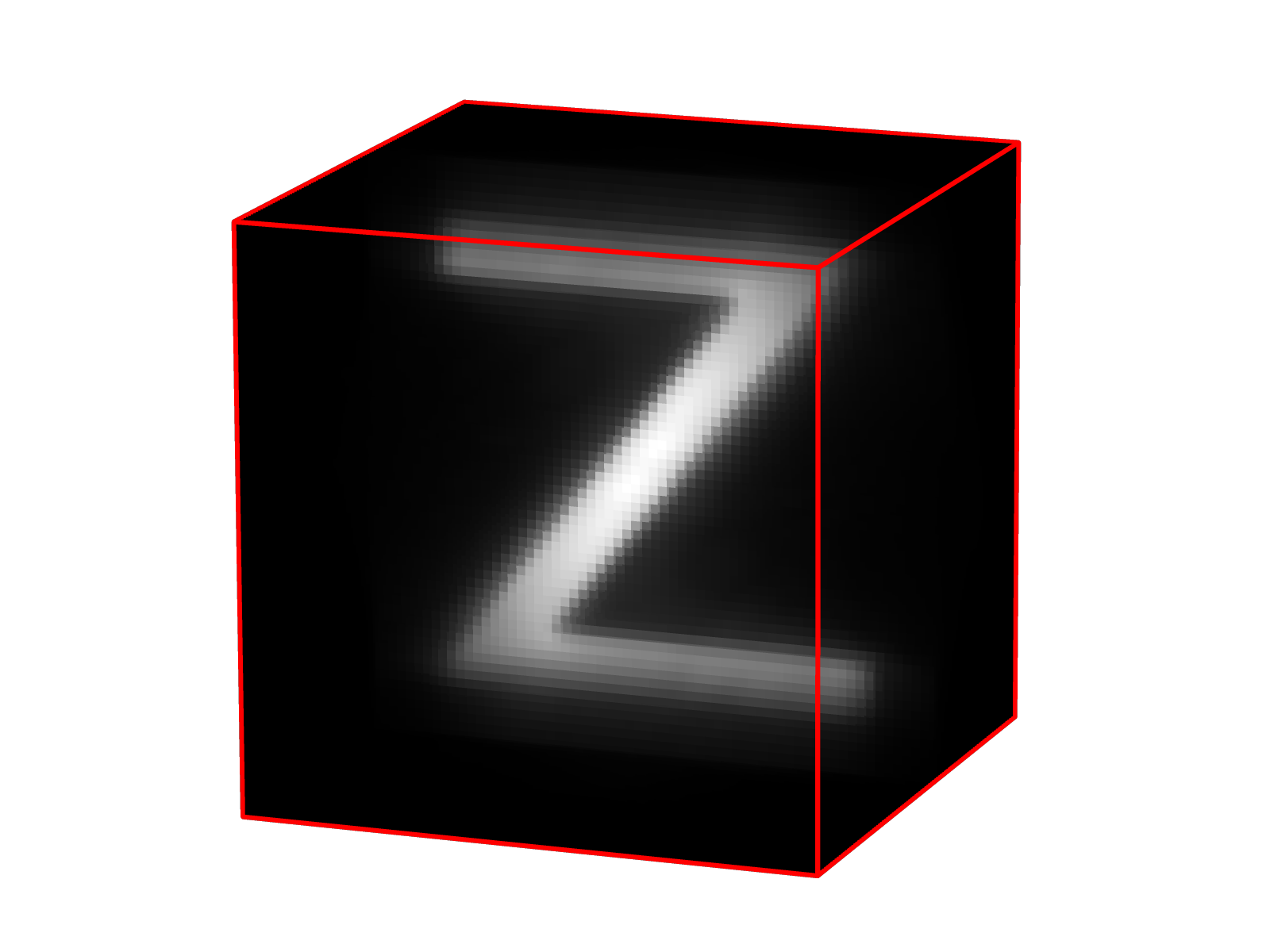}
	    \includegraphics[width=1\linewidth]{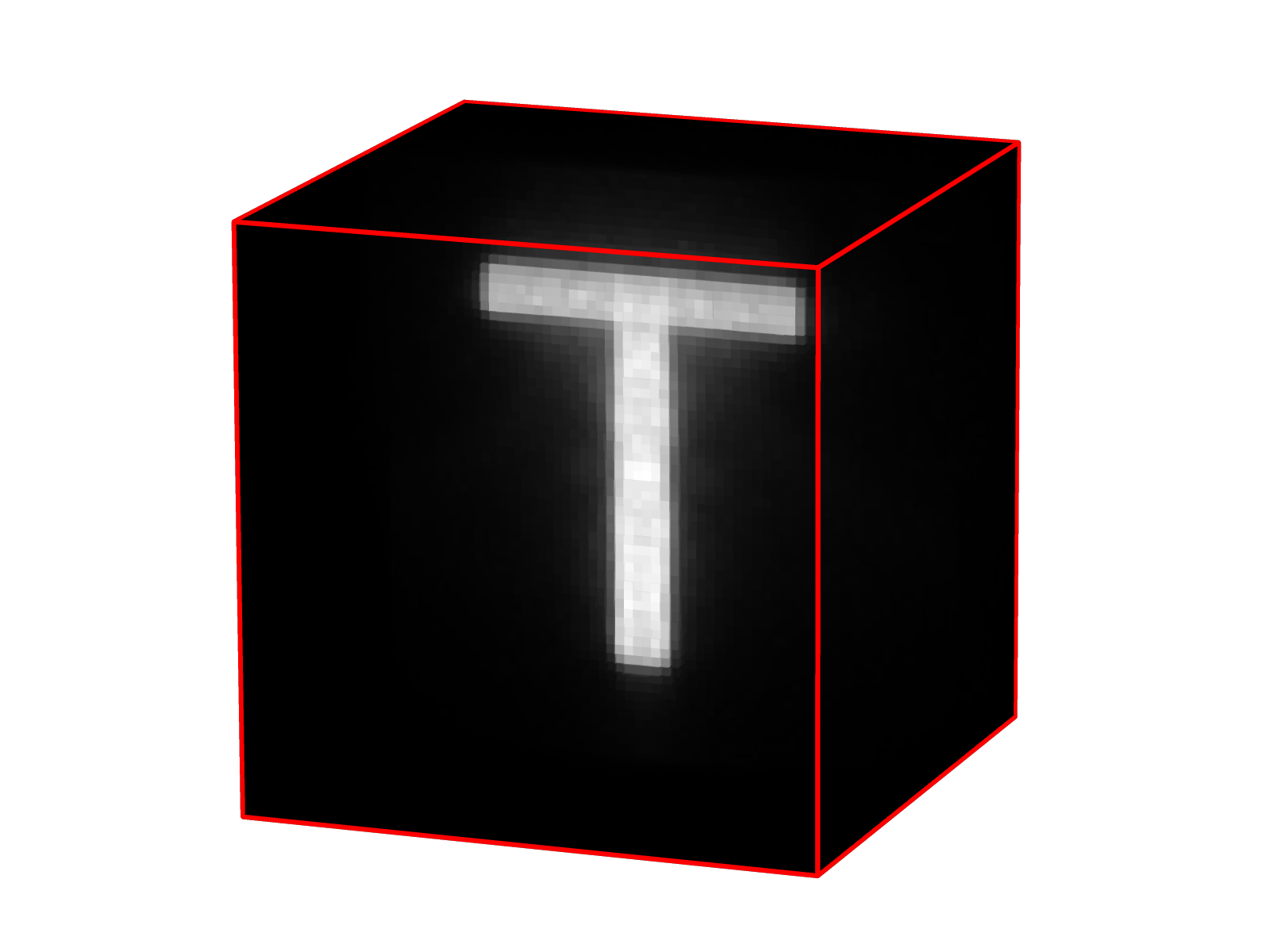}
		\caption{FK}
	\end{subfigure}
	\begin{subfigure}[]{0.32\linewidth}
		\centering
	    \includegraphics[width=1\linewidth]{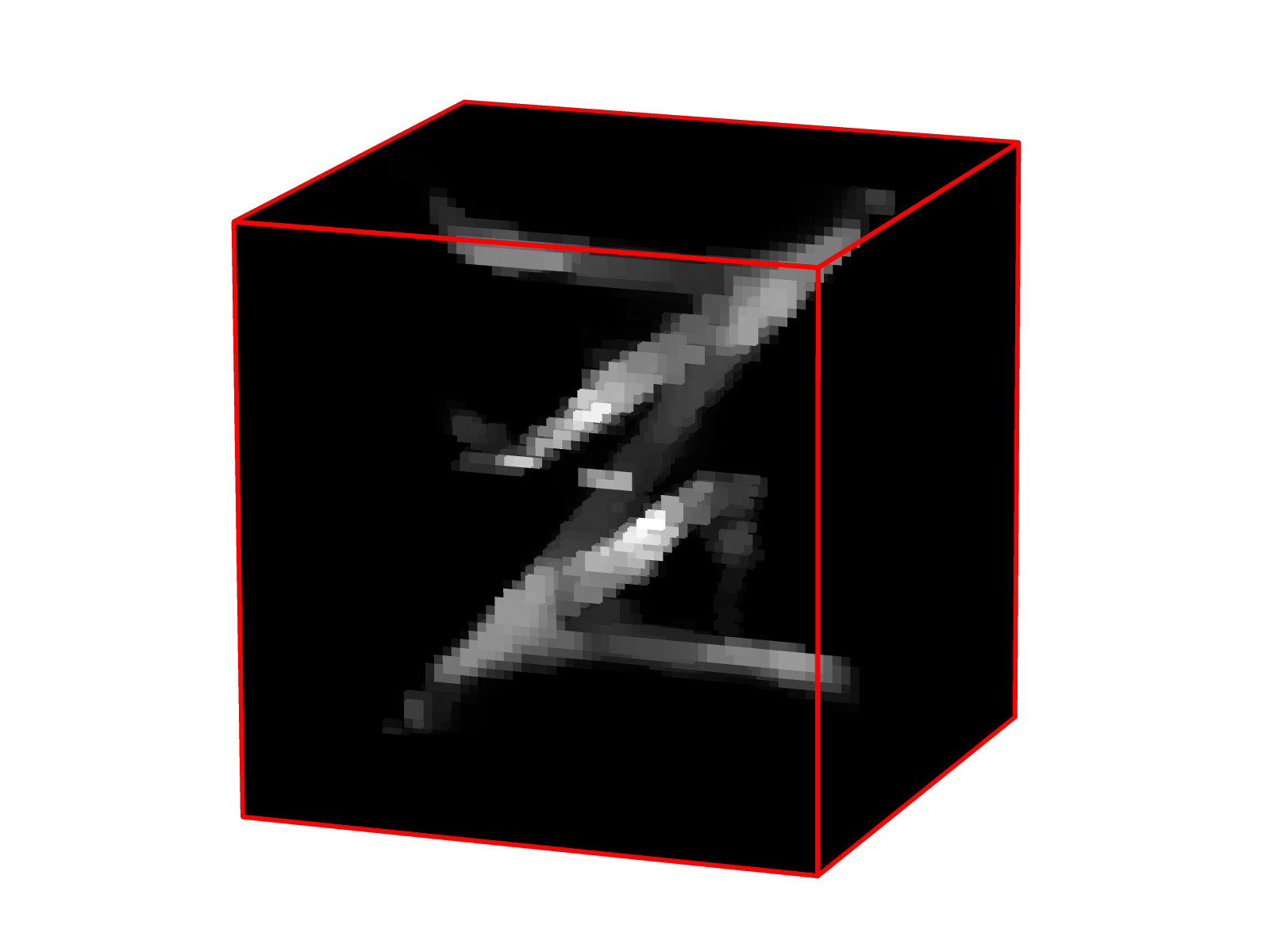}
	    \includegraphics[width=1\linewidth]{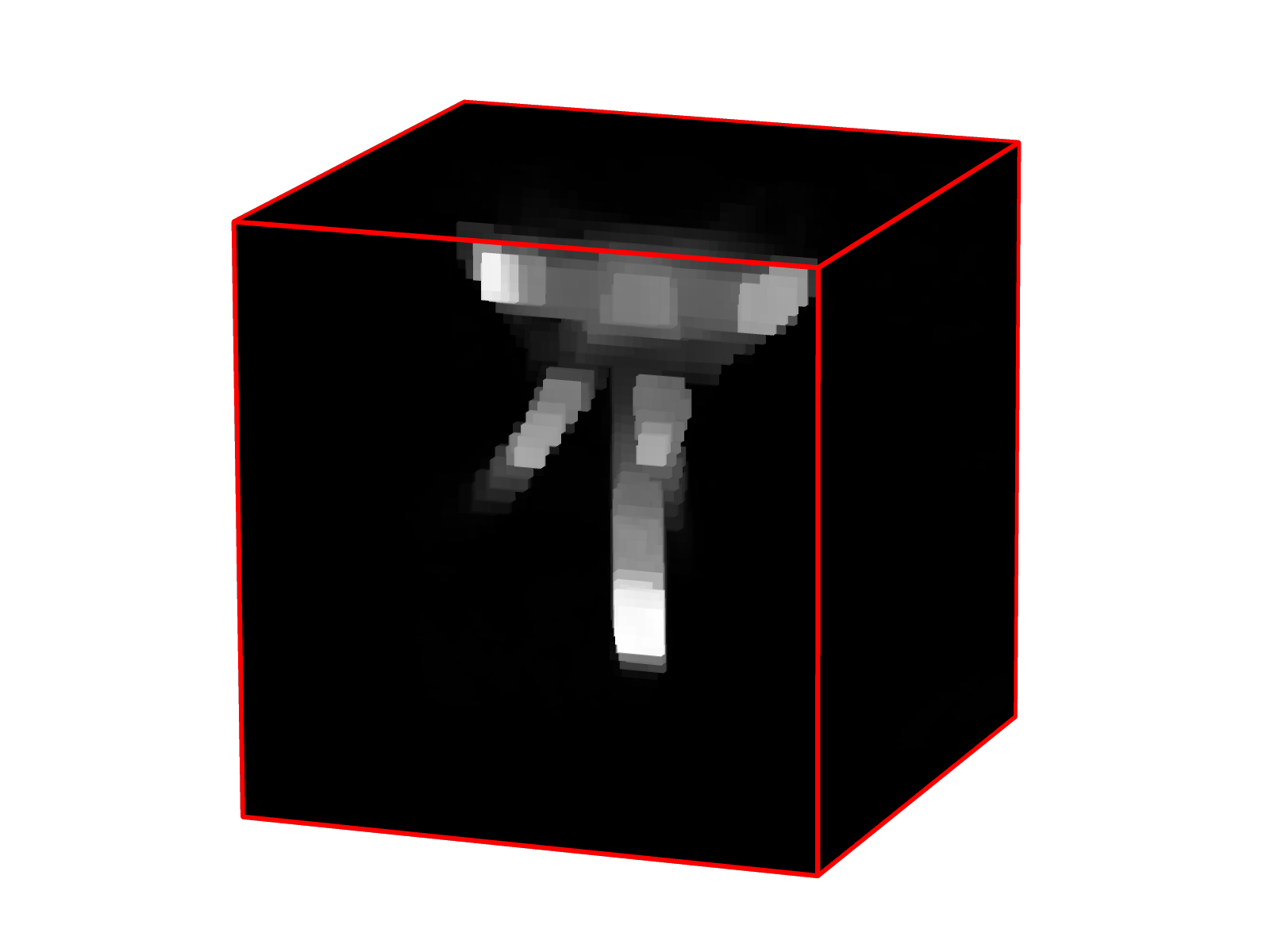}
		\caption{\CCNLOS}
	\end{subfigure}
	\caption{3D reconstruction examples on simulated data from the Z-NLOS Dataset~\cite{galindo19-NLOSDataset}. Our circular scans contain significant information about the hidden scene, but can potentially suffer from artifacting thanks to ambiguity in the measurements.}
	\label{fig:3d_sim}
\end{figure}

We show additional 3D reconstruction results in Fig.~\ref{fig:3d_real} and Fig.~\ref{fig:3d_sim}. In general, a \CCNLOS scan is sufficient for recovering the important shape of the hidden scene, as shown by our reconstructions. However, especially in the case of simulated data, not enough measurements are provided to resolve possible ambiguities in the voxel volume, resulting in streaking artifacts that degrade the output quality.

\vspace{-100mm}

\end{document}